\documentclass[journal,twocolumn,letterpaper]{IEEEJERM}

\ifCLASSINFOpdf
\else
\fi

\usepackage{times,amsmath,epsfig}
\usepackage{fancyhdr}
\usepackage{amsmath}
\usepackage{amsfonts}
\usepackage{amssymb}
\usepackage{array}
\usepackage{graphicx}
\usepackage{url}
\usepackage{subfigure}
\usepackage{bm}
\usepackage{xcolor}
\usepackage{soul}
\usepackage{amssymb}
\usepackage[breaklinks=true]{hyperref}
\usepackage{breakcites}
\usepackage{hyperref}
\usepackage{lipsum}
\usepackage{multirow, booktabs}
\usepackage{tabularray}
\usepackage{lscape}
\usepackage{bbm}
\usepackage{cite}
\usepackage{romannum}
\usepackage{adjustbox}
\usepackage{orcidlink}
\usepackage[official]{eurosym}

\usepackage{array}
\newcolumntype{x}[1]{>{\centering\arraybackslash\hspace{0pt}}p{#1}}

\usepackage{algorithm,algpseudocode}
\usepackage{amsmath}
\usepackage{amsfonts}
\usepackage{amssymb}
\usepackage{scalerel}
\usepackage{tikz}
\usetikzlibrary{svg.path}

\definecolor{orcidlogocol}{HTML}{A6CE39}
\tikzset{
  orcidlogo/.pic={
    \fill[orcidlogocol] svg{M256,128c0,70.7-57.3,128-128,128C57.3,256,0,198.7,0,128C0,57.3,57.3,0,128,0C198.7,0,256,57.3,256,128z};
    \fill[white] svg{M86.3,186.2H70.9V79.1h15.4v48.4V186.2z}
                 svg{M108.9,79.1h41.6c39.6,0,57,28.3,57,53.6c0,27.5-21.5,53.6-56.8,53.6h-41.8V79.1z M124.3,172.4h24.5c34.9,0,42.9-26.5,42.9-39.7c0-21.5-13.7-39.7-43.7-39.7h-23.7V172.4z}
                 svg{M88.7,56.8c0,5.5-4.5,10.1-10.1,10.1c-5.6,0-10.1-4.6-10.1-10.1c0-5.6,4.5-10.1,10.1-10.1C84.2,46.7,88.7,51.3,88.7,56.8z};
  }
}

\newcommand\orcidicon[1]{\href{https://orcid.org/#1}{\mbox{\scalerel*{
\begin{tikzpicture}[yscale=-1,transform shape]
\pic{orcidlogo};
\end{tikzpicture}
}{|}}}}

\begin{document}

%

\vspace{1 cm}

\title{Predicting Visit Cost of Obstructive Sleep Apnea 
using Electronic Healthcare Records with Transformer}

%
\author{Zhaoyang Chen~$^1$$^\dagger$ \orcidicon{https://orcid.org/0000-0003-2808-400X}, 
        Lina Siltala-Li~$^1$$^\dagger$ \orcidicon{https://orcid.org/0000-0001-8379-8764},
        Mikko Lassila$^1$,\\~
        Pekka Malo$^1$ \orcidicon{https://orcid.org/0000-0002-1583-015X},~ 
		Eeva Vilkkumaa$^1$ \orcidicon{https://orcid.org/0000-0002-4217-8697}~
	Tarja Saaresranta$^2$, \orcidicon{https://orcid.org/0000-0001-9328-963X}~
	Arho Veli Virkki$^2$ \orcidicon{https://orcid.org/0000-0001-7731-5869}
       
}


\twocolumn[
\begin{@twocolumnfalse}
  
\maketitle


\begin{abstract}
 
  Background: \textnormal{Obstructive sleep apnea (OSA) is growing increasingly prevalent in many countries as obesity rises. Sufficient, effective treatment of OSA entails high social and financial costs for healthcare.} Objective: \textnormal{For treatment purposes, predicting OSA patients' visit expenses for the coming year is crucial. Reliable estimates enable healthcare decision-makers to perform careful fiscal management and budget well for effective distribution of resources to hospitals. The challenges created by scarcity of high-quality patient data are exacerbated by the fact that just a third of those data from OSA patients can be used to train analytics models: only OSA patients with more than 365 days of follow-up are relevant for predicting a year's expenditures.} Methods and procedures: \textnormal{The authors propose a method applying two Transformer models, one for augmenting the input via data from shorter visit histories and the other predicting the costs by considering both the material thus enriched and cases with more than a year's follow-up.} Results: \textnormal{The two-model solution permits putting the limited body of OSA patient data to productive use. Relative to a single-Transformer solution using only a third of the high-quality patient data, the solution with two models improved the prediction performance's $R^{2}$ from 88.8\% to 97.5\%. Even using baseline models with the model-augmented data improved the $R^{2}$ considerably, from 61.6\% to 81.9\%.} Conclusion: \textnormal{The proposed method makes prediction with the most of the available high-quality data by carefully exploiting details, which are not directly relevant for answering the question of the next year's likely expenditure.} 

Clinical and Translational Impact Statement: \textnormal{Public Health-- Lack of high-quality source data hinders data-driven analytics-based research in healthcare. The paper presents a method that couples data augmentation and prediction in cases of scant healthcare data.}  \end{abstract}


\end{@twocolumnfalse}]


  \renewcommand{\thefootnote}{}
  
  \footnotetext[1]{The manuscript was submitted for review on xx.xx.2023.}
  \footnotetext[2]{This research was supported by the Foundation for Economic Education (grant number 16-9442), the Paulo Foundation, and the Helsinki School of Economics Foundation (HSE).}
  \footnotetext[3]{The main authors, denoted with $^\dagger$, are equal in their contributions.}
  \footnotetext[4]{The authors  denoted with $^1$ work with the Department of Information and Service Management, Aalto University, Finland. (correspondence email: lina.siltala-li@aalto.fi)} 
   
  \footnotetext[5]{The authors  denoted with $^2$ work with Division of Medicine, Department of Pulmonary Diseases, Turku University Hospital and Sleep Research Centre, Department of Pulmonary Diseases and Clinical Allegology, University of Turku.
}
 
%
\IEEEpeerreviewmaketitle

\section{Introduction}

\IEEEPARstart{O}{bstructive} sleep apnea (OSA) is a chronic respiratory disease in which the upper airway repeatedly collapses during sleep. There is no question that this results in poor sleep quality, thereby leading to increased daytime drowsiness, deterioration of cognitive abilities, various comorbidities, and even high rates of traffic and workplace accidents~\cite{walia2019beyond, garbarino2016risk}. In addition, many studies attest to high morbidity and mortality associated with the disease~\cite{bonsignore2019obstructive, chang2020obstructive,salman2020obstructive}. The prevalence of clinically diagnosed OSA was 3.7\% in the Finnish adult population~\cite{palomaki2022multimorbidity}. In Finland alone, 1.46 million people are estimated to suffer from moderate to severe sleep apnea, according to data presented in the Finnish Medical Journal~\cite{sleepapneainFinland2021}. This represents an astonishingly high percentage of the country's population of 5.54 million. In the wake of growing public awareness of the serious health consequences possible if OSA is left untreated, it is reported a significant increase in referrals connected with sleep apnea~\cite{sleepapneainFinland2019}.

It goes without saying that sufficient resources must be made available to match. All OSA patients should receive treatment. This requires physicians and healthcare decision-makers to plan budgets accurately and distribute supplies efficiently, for better resource allocation. Hence, they need information about the coming year's potential costs for OSA-related visits. While electronic healthcare records (EHRs) are ideal for training predictive models with data on visits to physicians, laboratory tests, and therapies, OSA complicates the use of this rich source of data because it is a chronic disease that involves irregular check-up intervals, extensive follow-up, and highly individualized treatments using evolving technologies. Therefore, a dragon of chaos exists in mining EHRs here, brought in by inconsistent coding over the years, large quantities of missing data, human input or measurement error, and loss of follow-ups. To at least some extent, these issues frequently arise in data analytics involving EHRs, which is unsurprising when one considers the messy landscape wrought by the complexity of pathology and epidemiology. Irrespective of these difficulties, the burgeoning quantities of data collected in EHRs renders them one of the best resources for healthcare research, and data-driven studies need to grapple with them~\cite{shah2020secondary}. Researchers take many approaches to the problem of EHRs' ``data chaos'' with one of the most popular applying state-of-the-art deep learning models since these do not presume any particular stochastic distributions to the data~\cite{choi2016doctor,choi2016retain,shickel2017deep,solares2020deep,kennedy2022augmentation}. Still, few studies address predicting the cost of healthcare visits in a way that accounts for both total costs and the visit type at each point in time, let alone focus on making the most of the limited body of data available for particularly complex diseases.

Our study represents five key contributions to the state of the art:

\romannum{1}) We develop a data-augmentation algorithm that preserves the semantic invariance of discrete healthcare data. 

\romannum{2}) We propose a method to augment the input via a subset of the high-quality healthcare data, material that cannot otherwise directly serve addressing the research question. 

\romannum{3}) A multi-task loss function is designed for cost prediction that considers both the sum-total costs and the cost specific to each type of visit. 

\romannum{4}) We combine two Transformer models (one for data augmentation and the other for cost prediction) to achieve better predictive performance while tackling the problem of insufficient data. 

\romannum{5}) Our research experiments with and hones the cost-prediction model by working with EHRs from Finland's Turku University Hospital. The study appears to be among very few projects of this type for OSA. The code from this study is available via \url{https://gitlab.com/lina.siltala/two_model_transformer_predict_cost}.

\section{Related work}
The worldwide volume of clinical data exceeded 2,300 exabytes in 2020~\cite{pramanik2022healthcare}. This vast body of data holds tremendous potential for data-driven analytics to support decision-making in healthcare, assessment of pathology trajectories, public-health surveillance, precision medicine, and preventive treatment, since EHRs encompass data covering consultations with experts, lab tests, clinical notes, and medication records~\cite{o2011impact,menachemi2011benefits,birkhead2015uses}. Since the sensitivity of the data held in EHRs makes them an obvious target for cyber-attacks and attractive for deliberate data leaks, strict regulations are in place for their use, such as the EU's General Data Protection Regulation (GDPR), the United States Health Insurance Portability and Accountability Act (or HIPAA), and rules for the Australian government's My Health Record system~\cite{shah2020secondary}. The ironic twist is that, through these, EHRs' treasure trove of data is not readily amenable to research. Scholars gain access to healthcare data only after a lengthy process for specified research questions, and the research must comply with ethics codes and rules -- e.g, the GDPR's terms for purpose limitation (Article 5(1)(b)), ``data minimization'' (Article 5(1)(c)), storage restrictions (Article 5(1)(e)), and integrity and confidentiality (Article 5(1)(f))~\cite{shah2020secondary}. It is, without doubt, imperative to protect individuals' privacy by means of standardization and strict ethics, yet this does bring challenges. For example, it is not easy to obtain the quantities of data needed for solid studies, especially with regard to particular diseases. This marks a stark contrast against natural language processing (NLP), for which Wikipedia, libraries, and social media offer ample material.

Several further factors contribute to the difficulties of scholarly use of EHRs. Firstly, the records are created primarily with physicians and administration in mind, not for research purposes~\cite{menachemi2011benefits}. Also, discrepancies arise, brought on by changes in technology, adjustments to diagnostic codes, and variations in practices between or even within healthcare facilities. Heterogeneous and free-form data create further difficulties for EHR analysis, as do complex intra-patient variations. A fourth important factor is that these records do not cover patients comprehensively: most people visit physicians only when unwell~\cite{shah2020secondary}. Of the many challenges bundled with EHRs' use in healthcare research, the two issues that we most needed to address for our study are the limited body of high-quality data available and the complicated characteristics of the data. 

Data augmentation is one technique for solving the first of these problems. Scholars of computer vision have frequently employed it for such purposes as cultivating more image data or applying clipping, rotation, color changes, or blurring. It is easy to understand how such techniques could serve such applications even without domain knowledge, since we know that the image is not converted to something completely different. That is, the post-augmentation body of data has retained the original's semantics~\cite{wei2019eda}. It is not so straightforward to apply these techniques to healthcare data. One of the reasons is that EHRs include many discrete variables. With these, keeping the semantic information intact is far more challenging than with the continuous variables that images involve.~\cite{lu2021textual} Work in the NLP domain, in contrast, points to possible ways forward, in that NLP data feature discrete variables and many scholarly efforts in that domain have tackled the problem of insufficient data, with a broad range of methods: random deletion, replacement, or injection~\cite{wei2019eda}; dependency-tree morphing~\cite{csahin2019data}; back-translation~\cite{feng2021survey}; the manifold mixup regularization method~\cite{verma2019manifold}; unsupervised data augmentation~\cite{xie2020unsupervised}; kernel methods~\cite{dao2019kernel}; semantic augmentation~\cite{nie2020named}; and others. Work specifically with data augmentation for EHRs has applied contrastive learning to find similarity patterns~\cite{wanyan2021bootstrapping}, examined particular types of data (such as images~\cite{wanyan2021bootstrapping} and textual clinical notes for patient-outcome prediction~\cite{lu2021textual}), and explored subfields such as skin-lesion analysis~\cite{perez2018data}. Scholars have discussed the potential for addressing the data-quantity issue with deep learning via knowledge distilling~\cite{che2015distilling}, patient representation~\cite{choi2016multi}, vector learning with non-negative restricted Boltzmann machines (eNRBMs)~\cite{tran2015learning}, and transfer learning in the EHR context~\cite{dubois2017effectiveness}. Workable data augmentation should be easy to implement while still improving the model's performance for the primary goal. All the aforementioned tactics turned out to be challenging to implement for our goal of predicting the next year's expenditures. While none of the techniques were directly applicable for our study, they offered inspiration for our augmentation method.

For each patient, the data records in our study are sequential and linked to specific visits, which vary in length. They are very similar to the many-to-many sequence-to-sequence (seq2seq) conditions in NLP~\cite{sutskever2014sequence}. Among the methods traditionally applied for seq2seq modeling are hidden Markov models (HMMs), latent semantic analysis (LSA), latent Dirichlet allocation (LDA), bag-of-words (BOW), skip-gram, words2vec, and global-vector representation~\cite{eddy1996hidden,lafferty2001conditional,blei2003latent,dumais2004latent,mikolov2013distributed,mikolov2013efficient,bojanowski2017enriching,pennington2014glove}. The renaissance of rapidly developing deep learning has channeled current approaches to sequential data mainly into the associated stream, though~\cite{iqbal2020survey}, with special attention surrounding the recurrent neural network and such variants as long short-term memory (LSTM)~\cite{salehinejad2017recent}. Among the state-of-the-art methods applied specifically in NLP-related work are variational autoencoders (VAEs)~\cite{kingma2013auto,semeniuta2017hybrid,yang2017improved,kingma2019introduction}, generative adversarial nets (GANs)~\cite{goodfellow2020generative,bowman2015generating}, adversarial learning for dialogue generation~\cite{li2017adversarial}, text generation with reinforcement learning~\cite{li2016deep, shi2018toward}, Transformer models~\cite{vaswani2017attention}, bidirectional encoder representation from Transformers (BERT)~\cite{devlin2018bert,rothe2020leveraging}, and individual solutions such as Generative Pre-trained Transformer 3 (GPT-2)~\cite{floridi2020gpt} and ChatGPT~\cite{chatgpt2022}. 

Following in the wake of the deep learning revolution in the NLP field, much research with EHRs has considered scalable deep learning in light of the two domains' parallels~\cite{shickel2017deep,rajkomar2018scalable,solares2020deep,esteva2019guide,miotto2018deep}. Many studies subject EHR data to deep learning for risk and disease prediction, data-privacy work, phenotyping, and disease classification, in a shift from labor-intensive feature engineering and other expert-driven methods. The target is data-driven approaches to representing complicated data in lower-dimensional space~\cite{xiao2018opportunities}. Several of these have demonstrated success in applying deep learning models in conjunction with EHRs. Among the tools produced are Deepr, using a conventional neural network (CNN) for deep extraction~\cite{nguyen2016mathtt}; DoctorAI, which utilizes a recurrent neural network (RNN) for disease prediction; and the DeepCare system, incorporating LSTM for predicting medicine quantities~\cite{pham2016deepcare}. Scholars have studied deep learning for numerous aspects of healthcare, such as predicting obesity~\cite{gupta2022obesity}, assessing the likelihood of readmission in cases of congestive heart failure~\cite{ashfaq2019readmission}, and providing diagnostic decision support via BERT~\cite{tang2021embedding}. Also, recent work has directed attention to the general issue of explainability, by means of RNNs and graphing of temporal data~\cite{choi2016doctor,choi2016retain,choi2017gram}.

Encoder--decoder was the most suitable state-of-the-art deep learning technique for our research. Encoders provide an embedding that can successfully learn the latent patient representation while converting the multivariates to a lower-dimensional space. Decoder architecture uses the latent representation discovered in the encoder phase as the context information to learn autoregressively about the following visit. Thus, the model takes age, gender, and other temporal multivariable features of historical patient data as inputs, while its output is temporally univariate (visit costs only). A Transformer model accommodates several inductive biases for sequential forecasting~\cite{tay2020efficient}. This model type offers one of the most powerful encoder--decoder architectures because it has multi-head attention and self-attention~\cite{vaswani2017attention}.

Transformers are frequently compared with CNNs and RNNs. A CNN induces the inductive biases of invariance and locality with kernel functions, while an RNN handles temporal-invariance and locality-related inductive biases via its Markovian structure. In contrast against both of these, Transformers do not demand any strong assumption as to the data's structural nature~\cite{tay2020efficient}. We refer the readers to the original paper of Transformer for more details~\cite{vaswani2017attention}. Although many studies apply Transformers accordingly~\cite{tay2020efficient,li2020behrt,lin2022survey}, few of them have produced EHR-based cost prediction that consider not only the sum of all costs but also the type of cost associated with each visit instance. Filling this gap, we took inspiration from the context-learning functionality of encoders that can retain semantic and syntactic information. Our literature-informed approach makes sure that the context of the patients is learned with a similar structure so that the patients' semantics do not change significantly during data augmentation. Applying two Transformers with the same encoder structure made it easy to implement data augmentation from a subset of the data.

\section{Material and Methodology}

Our data processing and methods are detailed below. The overall aim is to predict visit expenditures with a combination of original and augmented data. We address the data-augmentation and the cost-prediction element separately in relation to both the data and the model. 

\subsection{Data}

\begin{figure}[htbp]
    \centering
    \includegraphics[scale=0.2]{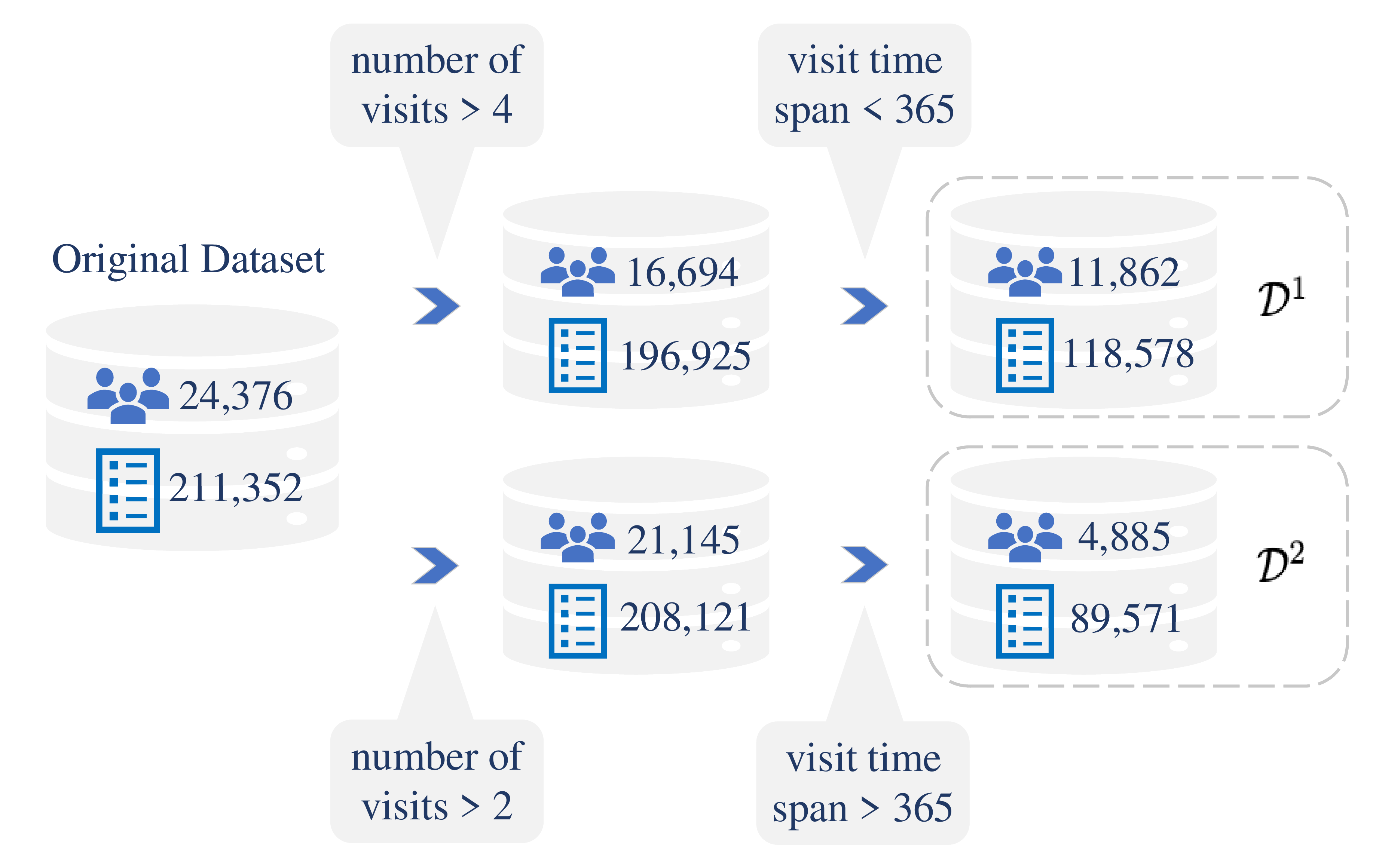}
    \caption{For analysis, the data were filtered by the number of visits and total follow-up duration. $\mathcal{D}^{1}$ is the set of patients with fewer than 365 days of follow-up, and $\mathcal{D}^{2}$ contains those patients with more than 365 days' follow-up. }
    \label{fig_filter_data}
\end{figure}

\begin{figure}[htbp]
    \centering
    \includegraphics[scale=0.17]{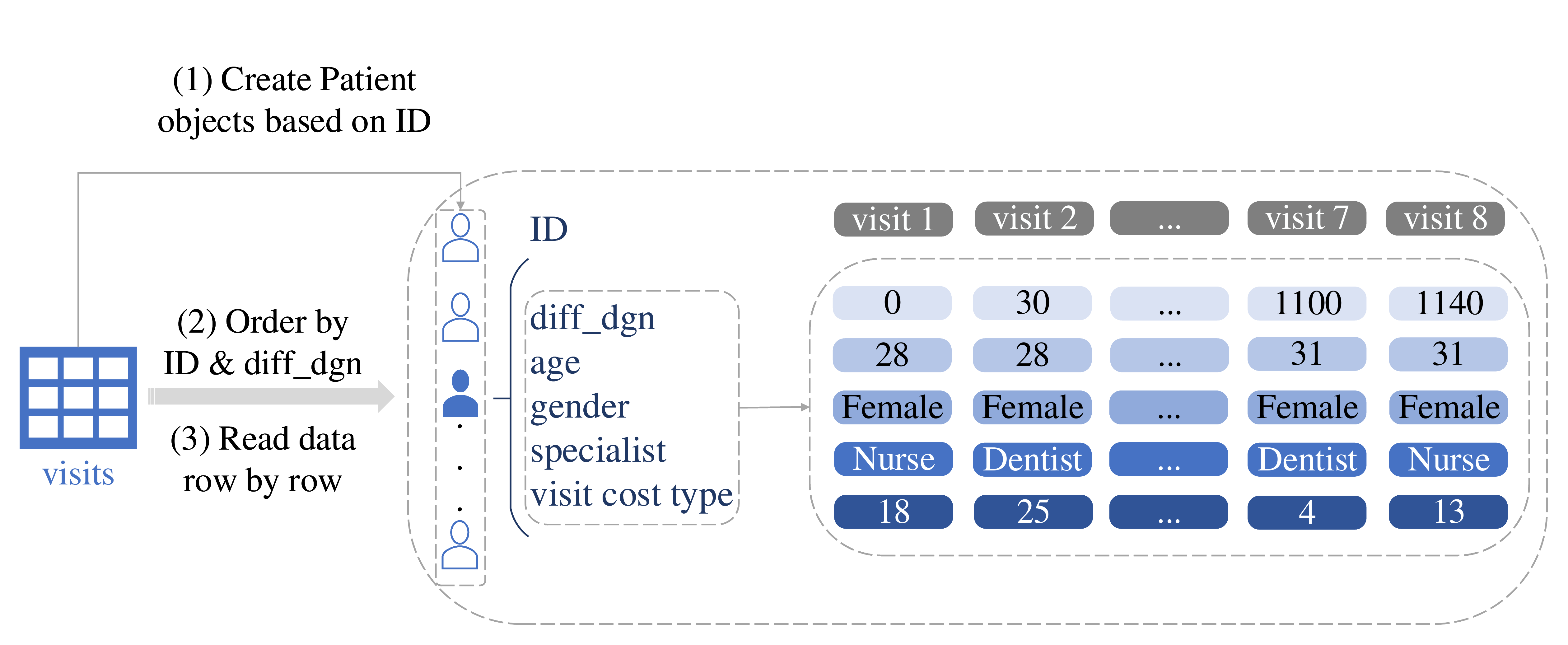}
    \caption{Variables for OSA patients. There are five patient-specific variables: days from diagnosis (\texttt{diff\_dgn}), age, gender, specialist type, and visit cost type.}
    \label{objectify}
\end{figure}

\begin{figure*}
\centering
    \includegraphics[width=0.4\textwidth]{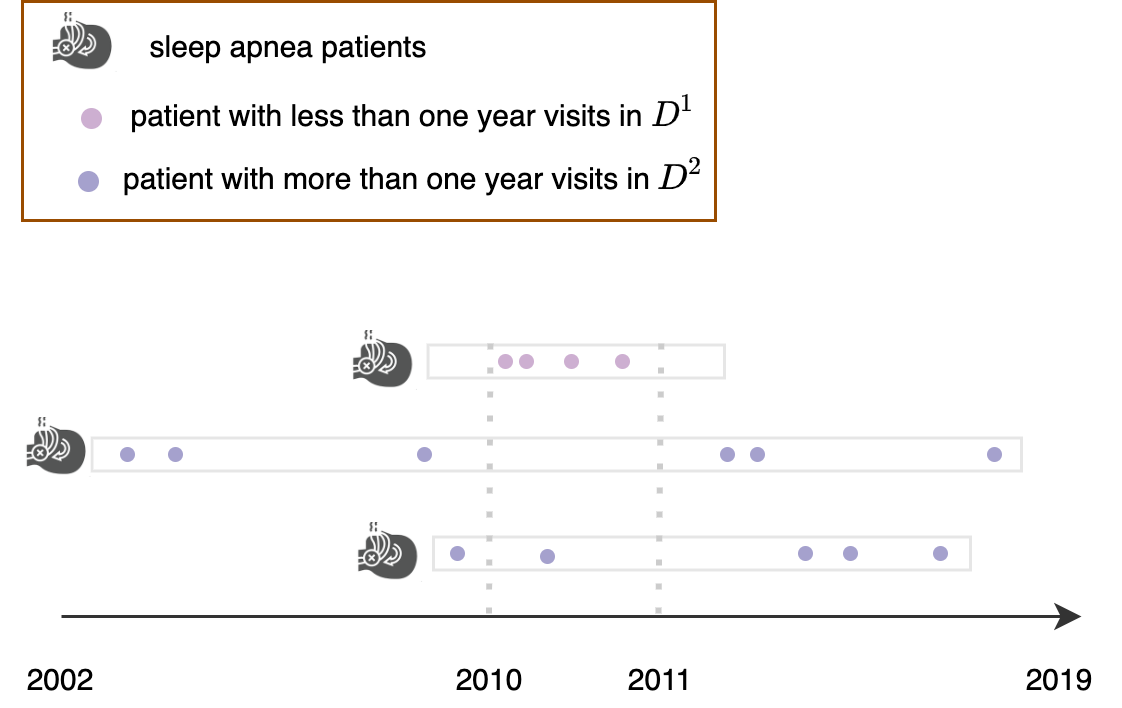}
    \includegraphics[width=0.4\textwidth]{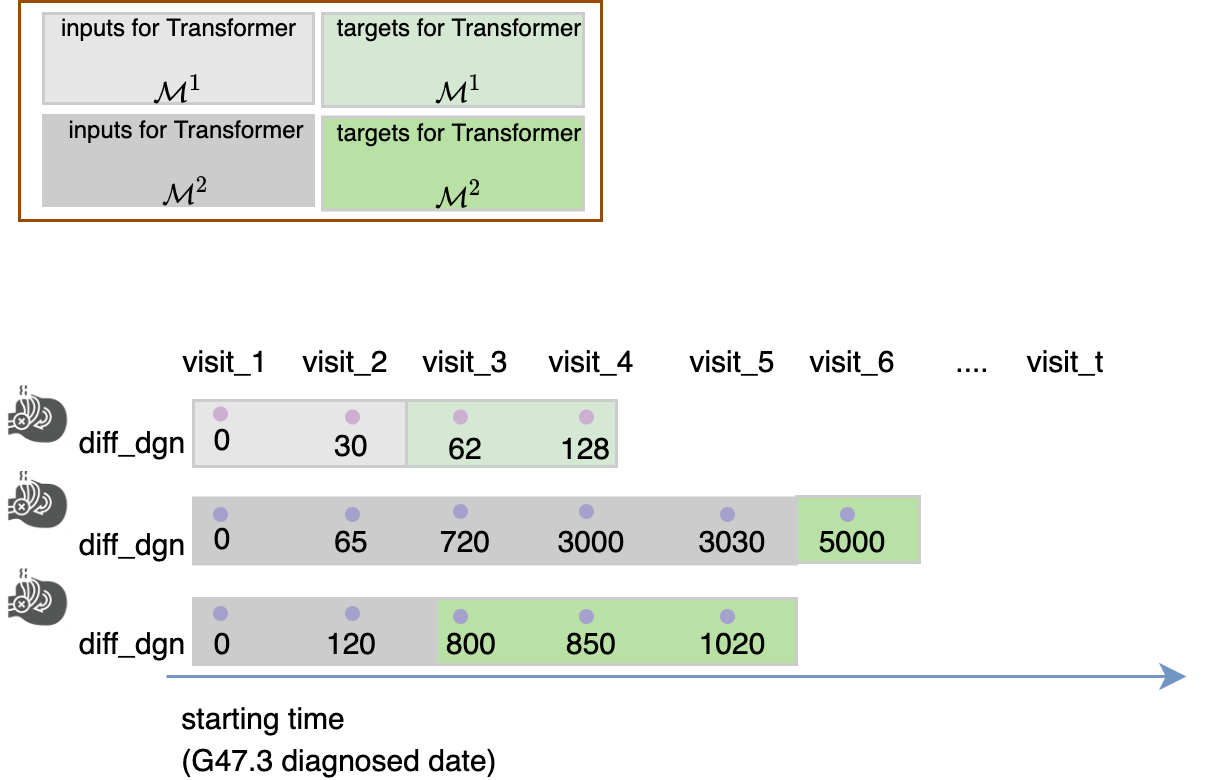}

\caption{Processing of the study's data. The figure illustrates the data-processing via three hypothetical patients, with (as shown at left) unique start dates, follow-up durations, inter-visit intervals, and visit frequency. Our processing used the date of G47.3 diagnosis as the start time. Each visit was processed as a discrete record with corresponding \texttt{diff\_dgn} values (shown in the right-hand pane). 
Patients with less than a year of follow-up were assigned to $\mathcal{D}^{1}$ while we placed the rest, with longer follow-up, in $\mathcal{D}^{2}$. The study used $\mathcal{D}^{1}$ to train Transformer $\mathcal{M}^{1}$ for data augmentation, while $\mathcal{D}^{2}$ was employed for visit cost prediction with Transformer $\mathcal{M}^{2}$.}
\label{fig_data_process}
\end{figure*}

\begin{figure*}
\centering
    \includegraphics[width=0.65\textwidth]{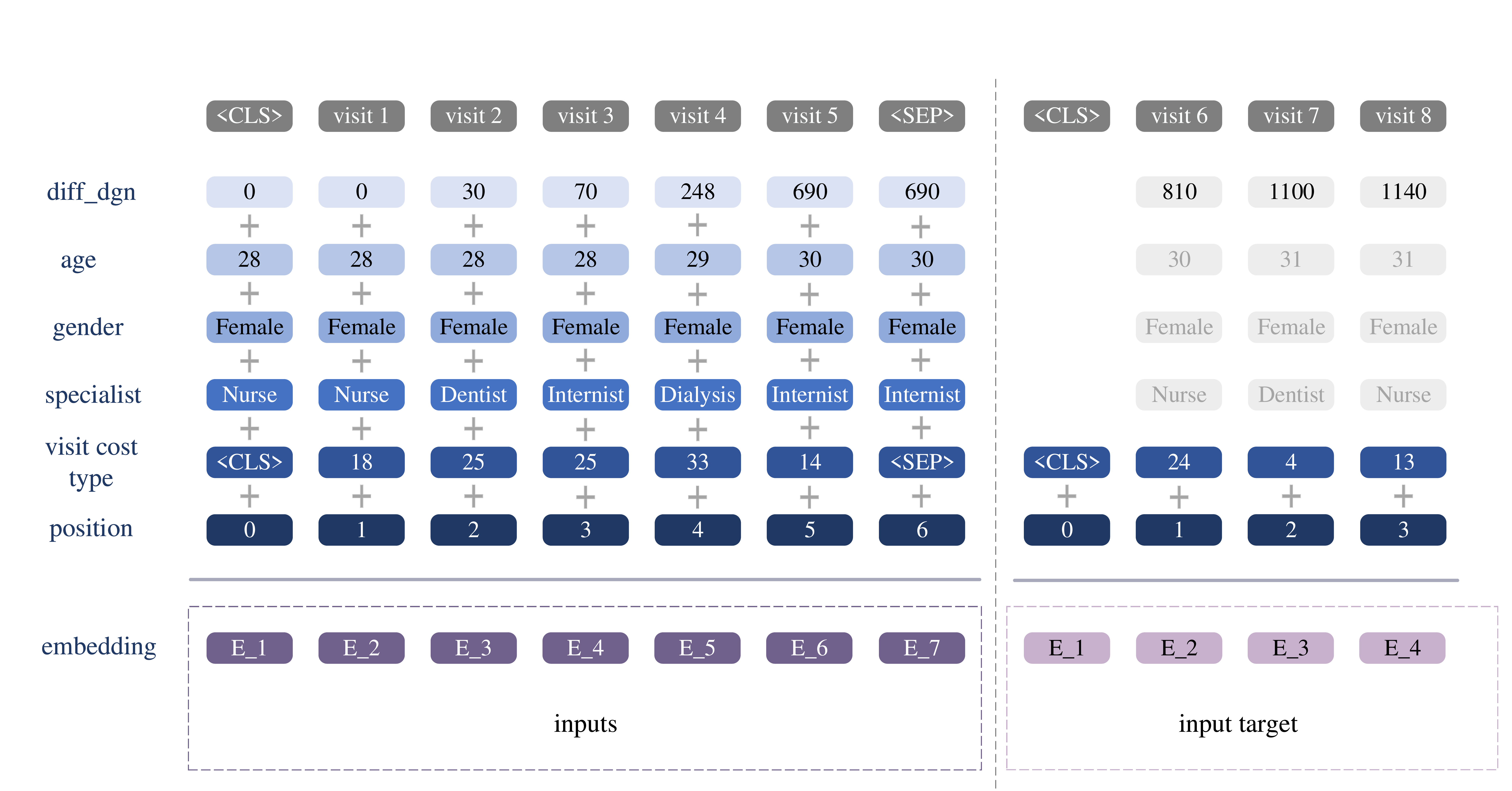}
    \includegraphics[width=0.3\textwidth]{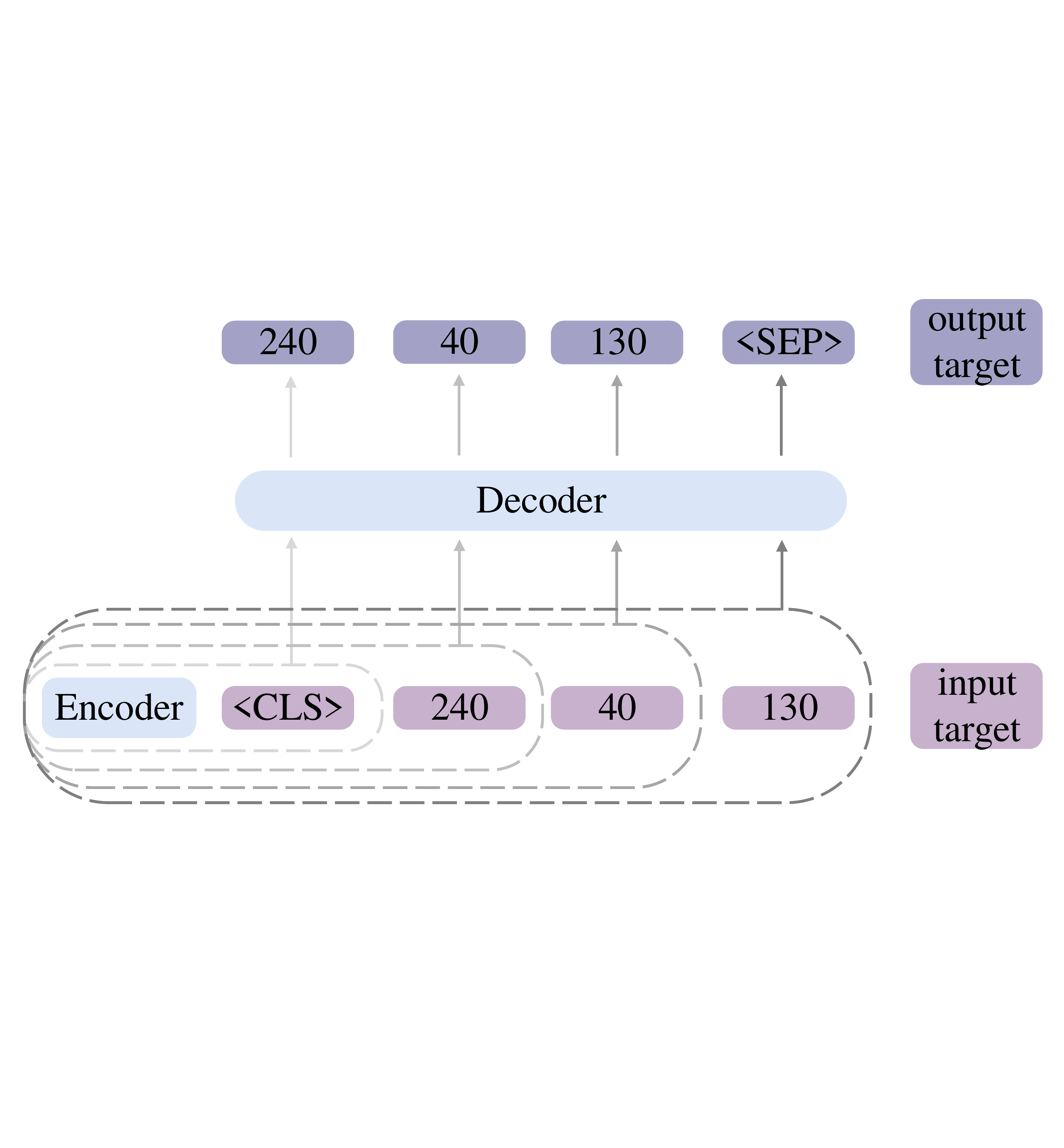}
\caption{Split the data into inputs and target, and inputs' embedding. The patient represented made eight visits, and the \texttt{diff\_dgn} set is $\{0,30,70,248,690,810,1100,1140\}$. We have records extending until 1,140 days after this patient's sleep-apnea diagnosis. The goal is to predict the visit cost over the last year ($1140-364 = 776$) on the basis of the previous visits' records (from the date of G47.3 diagnosis to day 776). Therefore, the time indices for the inputs and target are the maximum between 
0 and 776 and the minimum between 776 and the last visit point, day 1,140. In line with the definition $ t_{i}' = \max \{ \text{x} \in {T}_{i}: 0 \le \text{x} < t_{i} - 364 \} $ and $t_{i}'' = \min \{ \text{x}\in{T}_{i}:t_{i} - 364\le \text{x}\le t_{i}\}$, the inputs for this patient are the records from day 0 to day 690 from diagnosis, and the targets are the visit costs from day 810 to 1,140. The autoregressive prediction mechanism of Transformer is shown at right.}
\label{fig_data_process_embedding}
\end{figure*}

The filtering and preparation of data for this study are presented in Fig. \ref{fig_filter_data}, \ref{objectify}, and \ref{fig_data_process}. The data included the years from 2002 to 2019. Its procedures were approved under research permit T164/2019 from Turku University Hospital. On account of patients' irregular visit intervals and differences in follow-up duration, we would have had to contend with large quantities of ``missing data'' had we processed variables for all patients at the same time points, as studies often do~\cite{choi2016doctor,li2020behrt}. To circumvent the issue, we applied a data-processing ``trick'' from survival analysis~\cite{miller2011survival}: we set the date of the patient's first OSA diagnosis (identified as G47.3 in EHRs) as the start time of the study, then calculated the number of days between that and each visit for the variable \texttt{diff\_dgn}, for ``difference from diagnosis.'' Other variables were recorded for each visit instance alongside \texttt{diff\_dgn}. These captured both static and time-varying information, such as age, gender, the type of visit, and specialist type, as shown in Fig. \ref{objectify} and \ref{fig_data_process_embedding}. More detailed data descriptions and filtering of OSA cohort are presented in our supplementary material.

Formally, each patient $i$ is covered via sequential multivariate data until time ${t}_{i}$, where the value of ${t}$ at the maximum number of days from diagnosis may differ freely between patients on account of the differences in follow-up duration. All \texttt{diff\_dgn} values for patient $i$ form a  set ${T}_{i}=\{0,\dots,t_{i}, t_{i} \in 
 \mathbb{N}\}$. The elements of ${T}_{i}$ correspond to the patient’s visits, which can be represented by set ${S}_{i}=\{1, 2, 3, \dots,s_{i}, s_{i} \in 
 \mathbb{N}\}$, where $s_{i}$ is the total number of visits for patient $i$ (see Fig.~\ref{fig_data_process}). The patients' data are represented as $\mathcal{D} = \{\mathbf{d}_{ik}: i = 1, \dots,N,~ k \in {S}_{i} \}$, where ${N}$ is the number of patients and $\mathbf{d}_{ik}$ is a data vector with $M$ variables corresponding to patient $i$ and visit $k$. As Fig. \ref{fig_filter_data} and \ref{fig_data_process} indicate, we split the patients into two groups: those with under 365 days in all before the last visit ($\mathcal{D}^{1}$) and the other with more than 365 days' follow-up duration ($\mathcal{D}^{2}$).

\textbf{The data for augmentation $\mathcal{D}^{1}$}. The first set covers 11,862 patients, with 118,578 visit records. Because these patients' consultations spanned less than 365 days, they could not be included in the cost prediction for the next year, which requires more than one year of visits. Hence, only set $\mathcal{D}^{2}$ was available for cost prediction. That set contains only 4,885 patients. The low number of patients having data for more than 365 days is explained by the two facts: 1) about 30 \% of patients discontinue the treatment within the first year and 2) the number of patients has been much lower before the year 2019. Therefore, predicting a year's costs from so few patient data is highly challenging. It is natural to turn to data augmentation for a possible solution. In NLP settings, models that have learned the similarity of words can substitute another word for one that carries similar meaning in the context (e.g., ``cat'' in place of ``dog''), or some grammatical (semantic) information may be extracted such that deleting or inserting words yields a new sentence without distorting the meaning. healthcare-specific data augmentation, in contrast, is problematic, because such patterns of word similarity or grammar have not been found yet, especially with regard to certain diseases; therefore, there is no clear standard of what one can delete, replace, or inject for data augmentation that preserves the patient records' semantic information. Although we could not use $\mathcal{D}^{1}$ directly for the cost prediction, those sleep-apnea patients were treated at the same hospital as members of $\mathcal{D}^{2}$. Thanks to the associated similarities in sequence patterns and other characteristics, extracting information from $\mathcal{D}^{1}$ represented a feasible route for data augmentation to ameliorate the issue of the restricted pool of data for our prediction.

Data augmentation is designed to expand the input dataset in a manner that fills the material out with noise alongside the semantic information preserved, thereby improving the performance of the model~\cite{feng2021survey}. The augmentation in our case entailed changing only the details of one or two visits, so as to keep the characteristic sequence pattern of sleep-apnea patients intact. Hence, when we trained Transformer model $\mathcal{M}^{1}$ with $\mathcal{D}^{1}$, we processed the data as inputs and target in the following way: the patient history data $\mathcal{D}^{1}_{\text{inp}} = \{\mathbf{d}_{ik}:  i = 1, \dots, N_1, ~ 0 \le k \le s_{i}-2, ~ k\in {S}_{i} \}$, 
constitute the input, and the target (the patient's visit cost type) is $\mathcal{V}^{1}_{\text{out}} = \{\mathbf{v}_{ik}: i = 1, \dots, N_1,s_{i}-2 < k \le s_{i}, ~ k\in {S}_{i} \}$ as shown in Fig. \ref{fig_data_process}, where ${N}_1$ is the number of patients in $\mathcal{D}^{1}$,  and $\mathbf{v}_{ik}=[v_{ikc}]\in\{0,1\}^C$ is a binary vector, where $C$ is the number of unique visit cost types, and $v_{ikc}=1$ if and only if the cost type of visit $k$ for patient $i$ is $c$. 

\textbf{The data for cost prediction $\mathcal{D}^{2}$}. The 4,885-patient dataset contains 89,571 visit records. These data were processed as inputs and targets (per Fig. \ref{fig_data_process} and \ref{fig_data_process_embedding}) to predict, in our research setting, 
the visit costs over the last year (ignore leap year and only assume it has 365 days for simplicity). Since patients' visit history and intervals may be different in length, their time indices for the allocation of inputs and targets vary accordingly. We set $t_{i}' = \max \{ \text{x} \in {T}_{i}: 0 \le \text{x} < t_{i} - 364 \}$ as the last time for the inputs and $t_{i}'' = \min \{ \text{x}\in{T}_{i}:t_{i} - 364 \le \text{x}\le t_{i}\}$ as the first time for the targets. The visits corresponding to $t_{i}'$ and $t_{i}''$ are $s_{i}'$ and $s_{i}''$. Then, the inputs get expressed as $\mathcal{D}^{2}_{\text{inp}} = \{\mathbf{d}_{ik}:  i = 1, \dots ,N_{2}, 0 \le k \le s_{i}', k \in {S}_{i}\}$, and the targets (the patient's visit cost type) are $\mathcal{V}^{2}_{\text{out}} = \{\mathbf{v}_{ik}:  i = 1, \dots ,N_{2}, s_{i}'' \le k \le s_{i}, k\in {S}_{i} \}$. For simplicity, we denote the times corresponding to the inputs and targets as set ${T}_{\text{\textit{i}\textsubscript{\_inp}}} = \{0,\dots , t_{i}'\}$ and ${T}_{\text{\textit{i}\textsubscript{\_out}}} = \{t_{i}'',\dots , t_{i}\}$. 

Each unique visit cost type has an associated cost value. The patients' visit costs can be represented as cost vectors, $\mathcal{C}^{2}_{\text{out}} = \{\textbf{c}_{ik}:  i = 1, \dots,N_{2}, s_{i}'' \le k \le s_{i}, k\in {S}_{i} \}$, where the elements of $\textbf{c}_{ik}\in\mathbb{R}^C$ represent the costs attributed to different cost types. If the visit is not attributed to a cost type $c\in\{1,\dots,C\}$, then $c_{ikc}$ is set equal to zero. When feeding the data to the neural networks, special tokens $\big[\text{CLS}\big]$ and $\big[ \text{SEP}  \big]$ are inserted to indicate the first and final visit by the patient, respectively. 

It is worth noting that, because $\mathcal{D}^{1}$ contains many more patients than $\mathcal{D}^{2}$, and, consequently, also many more visit cost types (91 cost types in $\mathcal{D}^{1}$ versus 50 cost types in $\mathcal{D}^{2}$). To address the variation in the length of target sequences, all sequences are padded to have the same length.

\subsection{The model architecture}

\begin{figure*}[htbp]
    \centering
    \includegraphics[scale=0.2]{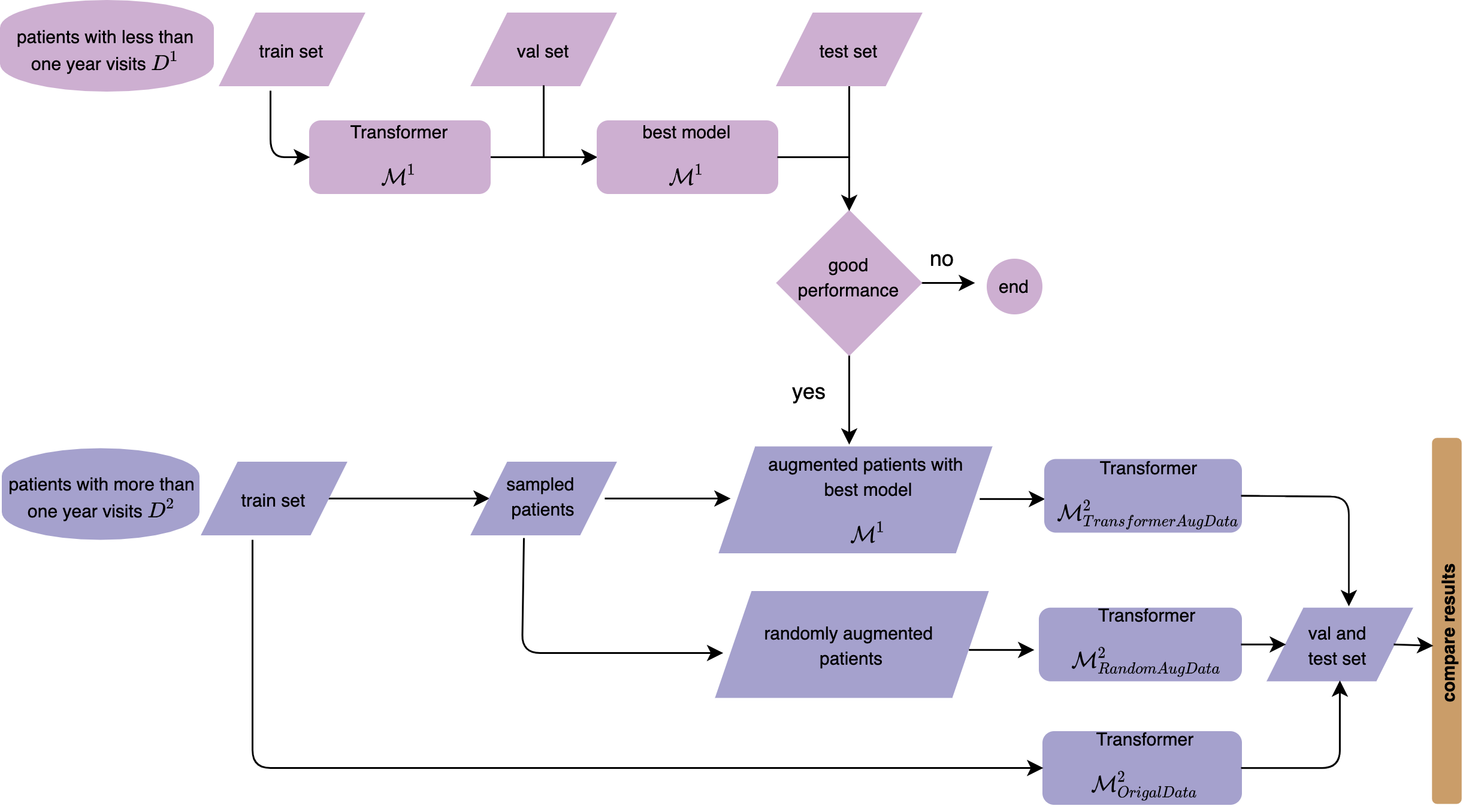}
    \caption{The model architecture. One of the two components is for data augmentation (applied with sub-sample $\mathcal{}D^{1}$), and the other is for cost prediction (utilizing augmented data and $\mathcal{}D^{2}$ in combination). The paper presents results from comparisons with the raw data, randomly augmented data, and Transformer $\mathcal{M}^{1}$ augmented data, to clarify the better prediction performance.}
    \label{fig_model_archi}
\end{figure*}

Encoder--decoder models are very popular for seq2seq prediction problems, and Transformers are among the most powerful tools in this class~\cite{lin2022survey}. Our choice to develop two Transformer models for enabling the efficient use of patient data in solid prediction of coming costs led to the architecture depicted in Fig. \ref{fig_model_archi}. This design addresses the two main differences between our work and application of seq2seq in NLP. 1) Rather than mostly univariate data (words in the NLP case), we had to factor in the multivariate nature of healthcare data (with variables such as age, gender, and specialists visited). For predicting next year's visit costs, historical cost information is not the only relevant variable. Others too are important, because they reflect between-patient differences and within-patient variance during follow-up -- demographic information plays important roles in visit patterns and trajectories. For our design to consider all of the most influential variables, we applied multivariable embedding in the encoder for $\mathcal{M}^{1}$ and $\mathcal{M}^{2}$, as shown in Fig. \ref{fig_data_process_embedding}. 2) Because of the small quantity of data for training and the challenges created for transfer learning by the complexity of pathology and epidemiology, we needed a solution for efficiently putting data to use in healthcare studies. The trained model $\mathcal{M}^{1}$ with $\mathcal{D}^{1}$ for data augmentation held promise to solve this two-horned problem 
via more data and an alternative solution for transfer learning.

\begin{figure*}[htbp]
    \centering
    \includegraphics[width=\textwidth]{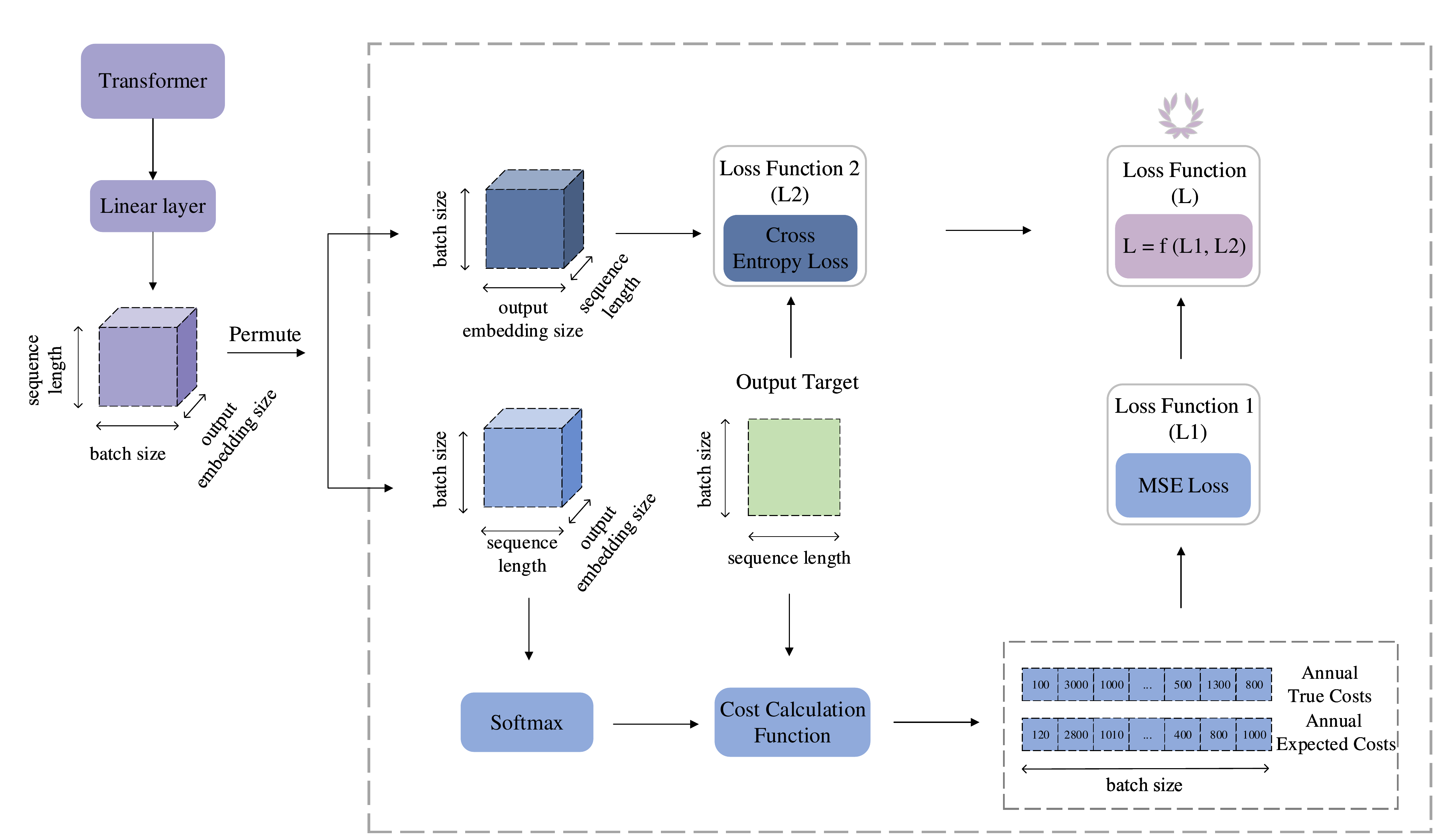}
    \caption{Loss function. To compute $\mathcal{L}_{1}$, we permute the dimension of the output of the linear layer, i.e., [sequence length, batch size, output embedding size], into [batch size, sequence length, output embedding size]. Then the output of the new dimension is passed to the Softmax layer to calculate the possibilities of the dimension of the output embedding size. Next, the cost calculation function helps calculate the labeled and predicted annual costs based on the output target and the Softmax output. The process is much simpler for calculating $\mathcal{L}_{2}$. The output of the linear layer is permuted into the dimension of [batch size, output embedding size, sequence length]. Then, the $cross\_entropy()$ function from PyTorch takes the permuted output and the output target to calculate the cross-entropy loss. Finally, since there is a large difference in magnitude between $\mathcal{L}_{1}$ and $\mathcal{L}_{2}$, we choose the common logarithm (log10) to scale down $\mathcal{L}_{1}$ and then add it to $\mathcal{L}_{2}$, i.e., $\mathcal{L}=log10(\mathcal{L}_{1}) + \mathcal{L}_{2}$.}
    \label{fig_model_enc_dec_loss}
\end{figure*}

\textbf{The model for data augmentation: $\mathcal{M}^{1}$}. As the diagram in Fig. \ref{fig_model_archi} indicates, $\mathcal{M}^{1}$ must be trained and evaluated before it gets employed for data augmentation. Although, as Fig. \ref{fig_model_enc_dec_loss} shows, $\mathcal{M}^{1}$ and $\mathcal{M}^{2}$ have identical encoder portions, the decoder in $\mathcal{M}^{1}$ uses only a fixed visit horizon (for augmentation inferring the next one or two visits) while the decoder part of $\mathcal{M}^{2}$ accounts for differences in visit length, since patients' visit frequencies in the next year will differ. 
The loss function in $\mathcal{M}^{1}$ measures cross-entropy, which is often used for multi-label classification. 

The algorithm developed for data augmentation includes deletion, replacement, and insertion, all of which are common tactics, especially for augmentation in situations with discrete variables~\cite{feng2021survey}. After training and evaluating $\mathcal{M}^{1}$ with $\mathcal{D}^{1}$, our method augments the data with randomly sampled patients $\mathcal{D}^{2}_{s}$ from $\mathcal{D}^{2}$ (line 1 in \verb|algorithm| \ref{alg:data_aug}). For our study, $\mathcal{M}^{1}$ was trained with $\mathcal{D}^{1}$, after which $\mathcal{M}^{1}$ is better suited to predicting the next visits for those patient having fewer than 365 days' follow-up as input data. For $\mathcal{D}^{2}$, with more than 365 days of follow-up, the algorithm specifies time indices ${T}^{2}_{s}$ that aid in extracting visits that are less than 365 days from patients 
in $\mathcal{D}^{2}_{s}$ (line 2), to guarantee input data suitable for inferences with $\mathcal{M}^{1}$. Visits up to ${T}^{2}_{s}$ are taken as inputs, and $\mathcal{M}^{1}$ infers the next one or two visits, predicting them from these inputs (lines 7-8). The next one or two visits predicted for these sample patients in light of their time indices are stored for later data augmentation. The data is stored for augmentation only if the prediction by $\mathcal{M}^1$ for the first visit is such that it is found in $\mathcal{V}^2$, which contains the true visit cost type vectors associated with patients in $\mathcal{D}^2$. If the predicted cost type is found only in $\mathcal{D}^1$ but not in $\mathcal{D}^2$, then the observation is omitted from the sample (lines 13-14).

The augmentation of predicted cost types depends on the accuracy of the prediction produced by $\mathcal{M}^1$. If the model is able to predict the first visit correctly, then the prediction for the second visit is augmented to the sample as a new observation (lines 16-17). However, if the prediction is incorrect for the first visit, but the visit cost type is still something that is found in $\mathcal{V}^2$, then the original visit cost type in the sample is still replaced with the predicted value even though the prediction is known to differ from the true value (lines 17-18). This is done to ensure that the entire augmented data sample is consistent with the prediction produced by $\mathcal{M}^1$. The predicted cost type for the second visit is omitted. 

In our experiment, the training set had 3,910 patients from $\mathcal{D}^{2}$ and the sampling ratio was set to 55\%, so the resulting set $\mathcal{D}^{2}_{s}$ included 2,150 patients. We iterated \verb|algorithm| \ref{alg:data_aug} three times and aggregated all results with the original training set, which gave us a total of 10,360 patients as $\mathcal{M}^{2}_{TransformerAugData}$ for $\mathcal{M}^{2}$ (as shown in Fig.~\ref{fig_model_archi}) for cost prediction.

\begin{algorithm}[htbp] 
\caption{-- Using Transformer $\mathcal{M}^{1}$ for data augmentation}
\label{alg:data_aug}
\begin{algorithmic}[1]
\Require{$\mathcal{M}_1$, $\mathcal{D}^{2}$, sample ratio} 
\vspace*{0.46\baselineskip}
\Ensure{\text{augmented data}}
\Statex

\vspace*{0.46\baselineskip}
\textbf{Step 1: Preparing the sample}
\vspace*{0.46\baselineskip}
\State Sample patients $\mathcal{D}^{2}_{s}$ from $\mathcal{D}^{2}$ with sample ratio 
\vspace*{0.46\baselineskip}
\State {Define ${T}^{2}_{s}$ as a set of visit indices in $\mathcal{D}^2_s$ that have taken place in less than 365 days}
\vspace*{0.46\baselineskip}
\State {Define $\mathcal{V}^2$ as the set of unique visit cost type vectors in $\mathcal{D}^2$}

\vspace*{0.46\baselineskip}
\State {Define $\mathcal{V}_s^2$ as the visit cost type vectors for the patients included in sample $\mathcal{D}_s^2$}

\vspace*{0.46\baselineskip}
\textbf{\newline Step 2: Predicting cost types of next visits in $\mathcal{D}^{2}_{s}$}
\vspace*{0.46\baselineskip}
\For{$\mathbf{d}_{ik} \in \mathcal{D}^2_s$} 
    \vspace*{0.46\baselineskip}
    
    \State \# Check that visit index can be used as input
    \vspace*{0.46\baselineskip}
    \State {\textbf{if} $k\in T_s^2$ \textbf{then}} 
    
    \hspace*{0.2pt} \# Predict the cost types of next visits
    
    \hspace*{0.2pt} $\hat{\mathbf{v}}_{ik}, \hat{\mathbf{v}}_{i(k+1)}  \gets \mathcal{M}^{1}\text{.predict}(\mathbf{d}_{ik})$
    \State {\textbf{else}} go to line 4
    \State {\textbf{end if}}

    \vspace*{0.46\baselineskip}
    \State \# Check that predicted visit cost is supported in $\mathcal{D}^2$
    \State \# Note: model $\mathcal{M}^1$ supports visit cost types that are 
    \State \# not found in $\mathcal{D}^2$
    
    \vspace*{0.46\baselineskip}
    \State {\textbf{if} $\hat{\mathbf{v}}_{ik}\notin \mathcal{V}^2$ \textbf{then}} 
    
    \hspace*{0.2pt} delete $\mathbf{d}_{ik}$ from $\mathcal{D}_s^2$

    \hspace*{0.2pt} go to line 4

    \State {\textbf{end if}}
    
    \vspace*{0.46\baselineskip}
    \State \# Augment predicted cost types to $\mathcal{D}_s^2$
    \vspace*{0.46\baselineskip}

    \State {\textbf{if} $(\hat{\mathbf{v}}_{ik}==\mathbf{v}_{ik})$ and $(\hat{\mathbf{v}}_{i(k+1)} \neq \text{null})$ \textbf{then}} 
    
    \hspace*{0.2pt} \# Case 1: next visit is correctly predicted by $\mathcal{M}^1$
    \vspace*{0.46\baselineskip}
    
    \hspace*{0.2pt} inject the 2nd predicted visit $\hat{\mathbf{v}}_{i(k+1)}$ in $\mathcal{D}^2_s$

    \hspace*{0.2pt} go to line 4

    \vspace*{0.46\baselineskip}
    \State {\textbf{else}}
    \vspace*{0.46\baselineskip}
    
    \hspace*{0.2pt} \# Case 2: next visit is predicted incorrectly but found 
    
    \hspace*{0.2pt} \# as a visit cost type supported by $\mathcal{D}^2$
        \vspace*{0.46\baselineskip}
        
    \hspace*{0.2pt} $\hat{\mathbf{p}}_{ik} \gets \mathcal{M}^1\text{.prob}(\mathbf{d}_{ik})$

    \hspace*{0.2pt} $c \gets \underset{c\in\{1,\dots,C\}}{\text{argmax}}\, \{\hat{p}_{ikc} \ : \ \exists \mathbf{v}_{ik}\in\mathcal{V}^2, v_{ikc}=1\}$

    \hspace*{0.2pt} define $\mathbf{v}$ such that $v_{c}=1$ and $v_{j}=0$ if $j\neq c$.

    \hspace*{0.2pt} replace $\mathbf{v}_{ik}$ with $\mathbf{v}$ in $\mathcal{D}_s^2$

    \State {\textbf{end if}}

\EndFor

\vspace*{0.46\baselineskip}
\end{algorithmic}
\end{algorithm}

\textbf{The model for cost prediction: $\mathcal{M}^{2}$}. Taking an approach similar to that in NLP, we compute the conditional probability of a patient's visit cost data $\mathcal{C}_{i}\subset \mathcal{C}^2_{\text{out}}$ for each patient $i$ given the corresponding input data $\mathcal{D}_{i}\subset\mathcal{D}^2_{\text{inp}}$ and previous visit costs

\begin{align}
\label{eq_cond_prob}
    &{P}(\mathcal{C}_{i} \mid \mathcal{D}_{i})  \nonumber\\
    & = {P}(\mathbf{c}_{i,s_{i}''} \mid \mathcal{D}_{i})\prod_{k=s_{i}''+1}^{s_i}{P}(\mathbf{c}_{i,k} \mid \mathbf{c}_{i,s_{i}''},\dots,\mathbf{c}_{i,k-1},\mathcal{D}_{i}).
\end{align}

\textbf{The loss function for $\mathcal{M}^{2}$.} Although our primary objective is to predict the sum-total visit costs for the next year on the basis of demographic details and information on visits (which occur sequentially during the last year of follow-up), 
it is important also to predict the type of visit cost at each time point. This kind of prediction is necessary for two reasons: 1) From a practical perspective, it is highly informative for healthcare decision-makers. With this information, they not only can calculate annual costs accurately for budget purposes but also can efficiently allocate specific resources to individual departments in accordance with the predicted visit cost types. 2) From the computation standpoint, predicting individual visits serves to regularize the cost prediction such that its performance can be improved by means of a regularizer when it is forced to consider both the total cost and distinct cost types at each time point. Therefore, the total-loss function of $\mathcal{M}^{2}$ is defined as $f(\mathcal{L}_{1}, \mathcal{L}_{2})$, which is a combination of regression cost $\mathcal{L}_{1}$ and multi-label cost $\mathcal{L}_{2}$. In our study, we have experimented with three different functions $f(\mathcal{L}_{1},\mathcal{L}_{2})$. Their results are presented in the supplementary material. Based on the results, we have selected $f(\mathcal{L}_{1}, \mathcal{L}_{2}) = \text{log}_{10}(\mathcal{L}_{1}) + \mathcal{L}_{2}$ as loss function.

Let $\mathcal{C}_b \subset \mathcal{C}^2_{\text{out}}$ be a minibatch of actual cost data for $N_b$ patients'. Let $\mathcal{D}_b\subset\mathcal{D}^2_{\text{inp}}$ be the corresponding sample from the input data set. The regression cost is given by
\begin{eqnarray}   \mathcal{L}_1(\mathcal{C}_b,\mathcal{D}_b)=\frac{1}{N_b}\sum_{i=1}^{N_b} \left(\sum_{k=1}^{K_i} \sum_{c=1}^C (\hat{p}_{ikc}e_c - {c}_{ikc})\right)^2
\end{eqnarray}
as the mean-squared error of the annual true cost and the annual predicted cost, where $K_i$ is the number of visits for patient $i$, $\hat{p}_{ikc}$ is the predicted probability that the cost type of the visit $k$ is $c$, and $e_c$ is the cost for a visit with type $c$. Since the model $\mathcal{M}^2$ does not directly predict the costs, we use the estimated probabilities $\hat{\mathbf{p}}_{ik}=\mathcal{M}^2\text{.prob}(\mathbf{d}_{ik})$, where $\mathbf{d}_{ik}$ is the data of patient $i$ associated with the visit $k$, together with expected visit type costs to approximate the total costs for a visit. Here, the actual cost of visit $k$ by patient $i$ is ${c}_{ikc}$, when the visit is known to be of cost type $c$. The predicted probabilities are calculated using soft-max function.

The multi-label cost $\mathcal{L}_2$ is defined as the cross entropy loss 
\begin{align}
    & \mathcal{L}_2(\mathcal{C}_b,\mathcal{D}_b) \nonumber \\
    & =-\frac{1}{N_b}\sum_{i=1}^{N_b}\left(\frac{1}{K_i}\sum_{k=1}^{K_i}\sum_{c=1}^C \log\frac{\exp(o_{ikc})}{\sum_{c=1}^C \exp(o_{ikc})}\right),
\end{align}
where $o_{ikc}$ are the unnormalized logits produced by the last linear layer of the model, $\mathbf{o}_{ik}=\mathcal{M}^2\text{.logits}(\mathbf{d}_{ik})$.

\subsection{Evaluation metrics}
Because our objectives encompassed predicting two distinct major elements -- the sum total of costs in the next year and each visit's cost type -- we used two sets of metrics for model evaluation: 1) for the regression modeling, root mean-squared error (RMSE) and $R^{2}$ to measure cost-prediction performance and 2) top-k accuracy indicators (\emph{k}=3,5,10) for evaluation of the classification performance.

\textbf{Metrics for classification ($\mathcal{M}^{1}$ and $\mathcal{M}^{2}$).} Recommendation systems' ability to find the best options is often judged in terms of top-k accuracy~\cite{cremonesi2010performance}. healthcare analytics work has often followed design philosophy with such a ``best bet'' concept because it reflects the mindset of physicians performing diagnosis as they assess which diseases could be considered and whether, upon examination, the culprit might indeed be one of the candidates found~\cite{choi2016doctor}. In calculation of top-\emph{k} accuracy, the prediction is deemed correct if the true label is among the model's \emph{k} prediction with the highest predicted likelihood~\cite{cremonesi2010performance}. 

The formula is presented in Eq. \ref{eq_topk}. To avoid confusion with our earlier notation, where $k$ is used as an index for the patient's visits, we will use $h$ instead of $k$ to denote the rank of the prediction, and we will use $H$ as the number of prediction allowed for every true label. Let $\hat{\mathbf{v}}_{ij}^h$ denote the predicted cost type vector for patient $i$ at visit number $j$, where the cost type is selected based on the $h$-th highest predicted likelihood. Let $\mathbf{v}_{ij}$ be the corresponding true label of visit cost type for patient $i$ at visit number $j$. Indicator function $\mathbbm{1}\left(\hat{\mathbf{v}}_{ij}^h = \textbf{v}_{ij}\right)$ has a value of 1 if $\hat{\mathbf{v}}_{ij}^h = \textbf{v}_{ij}$; otherwise, the value is 0. In our study, we took the values 3, 5, and 10 as \emph{h} for prediction of any single cost. 

 \begin{align}
    &\text{Top-\emph{k} accuracy}(\textbf{v},\hat{f} ) \nonumber \\
    & = \frac{1}{\sum_{i=1}^{N}({s}_{i}-s''_{i})}\sum_{i=1}^{N} \sum_{j = s''_{i}}^{s_{i}}\sum_{h=1}^{H} \mathbbm{1}\left(\hat{\mathbf{v}}_{ij}^h = \textbf{v}_{ij}\right)\label{eq_topk}
\end{align}

\textbf{Metrics for regression ($\mathcal{M}^{2}$).} For measuring how close the predicted cost $\hat{\mathcal{C}}^{2}_{\text{\textit{i}\textsubscript{\_out}}}$ is to the actual cost $\mathcal{C}^{2}_{\text{\textit{i}\textsubscript{\_out}}}$, we chose two commonly used metrics suited to evaluating regression models~\cite{draper1998applied}. In this evaluation, \emph{N} was the number of patients. 

 Firstly, RMSE (the square root of the mean-squared error) gave us the expected value of the squared error or loss and it was computed as the square root of $\mathcal{L}_{1}$. It enjoys widespread use because it is expressed in the same units as the response variable~\cite{draper1998applied}.

 $R^{2}$, in turn, expresses the proportion of the variance explained by the independent variables in the model~\cite{botchkarev2018performance}. Via the proportion of the variance explained, it shows how well the model can predict the unseen data. The best possible $R^{2}$ value is 1.0, while a value of 0.0 indicates that the model does not aid in explanation (i.e., it predicts the average value $\bar{\text{c}}_{ikc}$).\cite{draper1998applied}  Eq. \ref{eq_R2} presents the calculation of $R^{2}$, where $\hat{\text{c}}_{ikc} = \hat{p}_{ikc}e_c$:
 \begin{equation}
    \textit{R}^{2} = 1- \frac{\sum_{i=1}^{N_b} \left(\sum_{k=1}^{K_i} \sum_{c=1}^C (\hat{{c}}_{ikc} - {c}_{ikc})\right)^2}{\sum_{i=1}^{N_b} \left(\sum_{k=1}^{K_i} \sum_{c=1}^C ( \bar{{c}}_{ikc} - {c}_{ikc})\right)^2}
    \label{eq_R2}
\end{equation}

\section{Results}

Our design split the dataset into training, validation, and testing sets. The purpose of validation is to prevent overfitting during training and to guarantee that the model gets evaluated in relation to an entirely unseen set of data~\cite{ottom2022znet}. The batch size and the learning rate were set as 64 and 0.0001 respectively. More information on experiments of hyperparameters can be found in the supplementary material. The performance of model $\mathcal{M}^{1}$ is presented in Table \ref{tab:model1_performance}. Since all the top-$\emph{k}$ values for the test set exceed 91\%, we apply the train $\mathcal{M}^{1}$ model for data augmentation based on the model design architecture shown in Fig.~\ref{fig_model_archi}.

\begin{table}[htbp]
    \centering
    \caption{Top-$\emph{k}$ performance of $\mathcal{M}^{1}$.}
    \begin{tabular}{x{1.7cm} x{1.7cm} x{1.7cm} x{1.7cm}} 
     \toprule
     \textbf{Set} & \textbf{Top3} & \textbf{Top5} & \textbf{Top10} \\ [0.5ex] 
     \hline\hline
     Train & 95.92\% & 97.70\% & 99.08\% \\ 
     \hline
     Val & 87.15\% & 90.77\% & 94.68\% \\
     \hline
     Test & 91.26\% & 94.04\% & 96.64\% \\
     \bottomrule
    \end{tabular}
    
    \label{tab:model1_performance}
\end{table}

\begin{table*}[bt!]
  \caption{Models performance with original data and augmented data with Transformer.}
  \label{models_results}
  \centering

  \begin{adjustbox}{width=\textwidth}
  \begin{tabular}{llccccccccccccccc}

    \toprule
    \multicolumn{1}{l}{} & \multicolumn{1}{l}{} & \multicolumn{5}{c}{\bfseries Original Data} &
    \multicolumn{5}{c}{\bfseries Randomly Augmented Data} & 
    \multicolumn{5}{c}{\bfseries Augmented Data with Transformer$\mathcal{M}^{1}$}  \\
     \cmidrule(rl){3-7}\cmidrule(rl){8-12}\cmidrule(rl){13-17}
    \textbf{Model} & {Set} & {Top3} & {Top5} & {Top10} & {RMSE} & {$R^{2}$} 
    & {Top3} & {Top5}  & {Top10} & {RMSE} & {$R^{2}$}
    & {Top3} & {Top5}  & {Top10} & {RMSE} & {$R^{2}$}\\

    \midrule\midrule[.1em]

    \multirow{3}{1.1cm}{LSTM \\ without attention} 
    & Train & 57.87\% & 73.32\% & 80.32\% & 403.63 & -0.002 
    & 63.90\% & 73.75\% & 80.55\% & 183.18& 0.590 
    & 61.62\% & 73.34\% & 80.13\% & 347.08 & 0.317 \\
    
    & Val   & 57.92\% & 72.82\% & 80.04\% & 419.22 & -0.001 
    & 62.90\% & 73.00\% & 80.09\% & 354.92 & 0.124
    & 62.35\% & 74.30\% & 80.74\% & 273.83 & 0.152 \\
    
    & Test  & 60.46\% & 75.48\% & 80.85\% & 297.77 & -0.011 
    & 66.00\% & 76.09\% & 81.22\% & 312.43 & -0.098
    & 62.08\% & 73.65\% & 80.28\% & 353.81 & 0.036\\

    \midrule
    
    \multirow{3}{1.1cm}{LSTM \\ with \\attention} 
    & Train & 60.72\% & 74.02\% & 80.33\% & 121.19 & 0.910 
    & 64.19\% & 76.44\% & 80.80\% & 93.86 & 0.943
    & 60.95\% & 76.32\% & 80.92\% & 103.04 & 0.941 \\
    
    & Val   & 60.96\% & 73.36\% & 80.10\% &  222.66 & 0.728
    & 65.43\% & 75.88\% & 80.09\% & 175.20  & 0.833
    & 64.50\% & 76.54\% & 80.70\% & 96.62 & 0.890 \\
    
    & Test  & 60.46\% & 75.48\% & 80.85\% & 195.85 & \textbf{0.884}
    & 64.70\% & 76.34\% & 80.55\% & 161.64 & 0.697
    & 63.49\% & 75.20\% & 79.85\% & 133.59 & \textbf{0.858} \\
    
    \midrule
    
    \multirow{3}{1.1cm}{BiLSTM \\ without \\attention} 
    & Train & 62.95\% & 73.40\% & 80.10\% & 183.18 & 0.794
    & 66.98\% & 76.54\% & 81.08\% & 135.95 & 0.881
    & 66.65\% & 75.92\% & 80.87\% & 94.51 & 0.951 \\
    
    & Val   & 59.07\% & 72.53\% & 79.57\% & 418.73 & 0.036
    & 62.94\% & 75.38\% & 79.96\% & 340.36 & 0.094
    & 62.93\% & 76.38\% & 80.76\% & 294.24 & -0.011 \\
    
    & Test  & 60.54\% & 75.74\% & 80.89\% & 311.16 & -0.074
    & 64.73\% & 76.82\% & 81.39\% & 290.28 & -0.020
    & 62.77\% & 75.67\% & 79.78\% & 363.77 & 0.011 \\
    
    \midrule
    
    \multirow{3}{1.1cm}{BiLSTM \\ with \\attention} 
    & Train & 61.26\% & 74.10\% & 80.40\% & 118.44 & 0.914
    & 65.34\% & 77.24\% & 81.31\% & 88.53 & 0.950
    & 65.73\% & 77-11\% & 81.26\% & 92.21 & 0.953 \\
    
    & Val   & 59.74\% & 72.15\% & 79.85\% & 242.76 & 0.678
    & 64.16\% & 76.21\% & 80.79\% & 223.40& 0.732
    & 65.35\% & 77.84\% & 81.00\% & 100.02 & 0.878 \\
    
    & Test  & 63.08\% & 76.29\% & 80.94\% & 175.97 & 0.616
    & 64.56\% & 77.56\% & 81.34\% & 183.75& 0.624
    & 64.33\% & 76.58\% & 80.84\% & 151.35 & \textbf{0.819} \\
    
    \midrule
    
    \multirow{3}{1.4cm}{Transformer $\mathcal{M}^{2}$} 
    & Train & 80.62\% & 86.30\% & 92.46\% & 29.63 & 0.994
    & 78.72\% & 85.41\% & 93.60\% &  8.96 & 0.999
    & 78.22\% & 85.06\% & 92.88\% & 46.15 & 0.988 \\
    
    & Val   & 81.36\% & 86.27\% & 93.31\% & 33.84 & 0.995
    & 80.61\% & 87.62\% & 93.56\% & 2.29 & 1.000
    & 79.85\% & 87.25\% & 93.22\% & 21.64 & 0.995 \\
    
    & Test  & 83.64\% & 87.76\% & 93.45\% & 116.89 & \textbf{0.888}
    & 78.00\% & 83.58\% & 92.98\% & 94.51  & \textbf{0.944}
    & 80.25\% & 86.12\% & 93.14\% & 57.89 & \textbf{0.975} \\
    
    \bottomrule

  \end{tabular}
\end{adjustbox}
\end{table*}

For the baseline models required as a reference for judging our two-model design, we have chosen four seq2seq models based on LSTM because LSTM is commonly used for seq2seq as one of the most popular encoder-decoder models. These four baseline models are: 1) a general LSTM-based encoder--decoder model, 2) an attention-oriented one, 3) a bidirectional one, and 4) an attention-based bidirectional one. Taking a combination approach, the last of these encoder--decoder models employs a bidirectional LSTM encoder to encode the historical time series while another LSTM decoder produces the future time series, with an attention mechanism implemented for coordinating the input and output time series and dynamically selecting the most pertinent contextual data for prediction purposes~\cite{yuan2020using}.
The other three baseline models are variants of the attention-based bidirectional LSTM (BiLSTM) encoder--decoder model. 
All four baseline models can learn from historical records and, thereby, create representative material to inform prediction.

We compared these models with our Transformer-based prediction, assessing the performance of each with the original raw data, randomly augmented data, and data augmented via Transformer model $\mathcal{M}^{1}$. This section details the results, presented concisely in Table \ref{models_results}. Since our goal entailed giving the highest priority to predicting regression cost ($\mathcal{L}_{1}$), special emphasis is placed on regression metrics ($R^{2}$). 


As the table attests, models without an attention mechanism display excellent top-\emph{k} accuracy but have negative $R^{2}$ values, suggesting that these models fail to predict the year's total visit costs. In contrast, LSTM with attention and our $\mathcal{M}^{2}$ achieves an $R^{2}$ of more than 0.88. Transformer $\mathcal{M}^{2}$ stands out from all the other models for every indicator. As for performance with randomly augmented data for training and evaluation, the LSTM models do worse, BiLSTM models function slightly better, and -- surprisingly -- $\mathcal{M}^{2}$ has a higher $R^{2}$ value: 0.944. Finally, when compared to the original data, data augmented via the same Transformer model ($\mathcal{M}^{2}$) afford better prediction of cost for all regression-model conditions except LSTM with attention (though the latter still is able to reach an $R^{2}$ of 0.858). The $R^{2}$ of Transformer $\mathcal{M}^{2}$ rises to 0.975 in this condition, and that of BiLSTM increases from 0.616 to 0.819. 

\begin{figure}[htbp]
    \centering
    \includegraphics[scale=0.35]{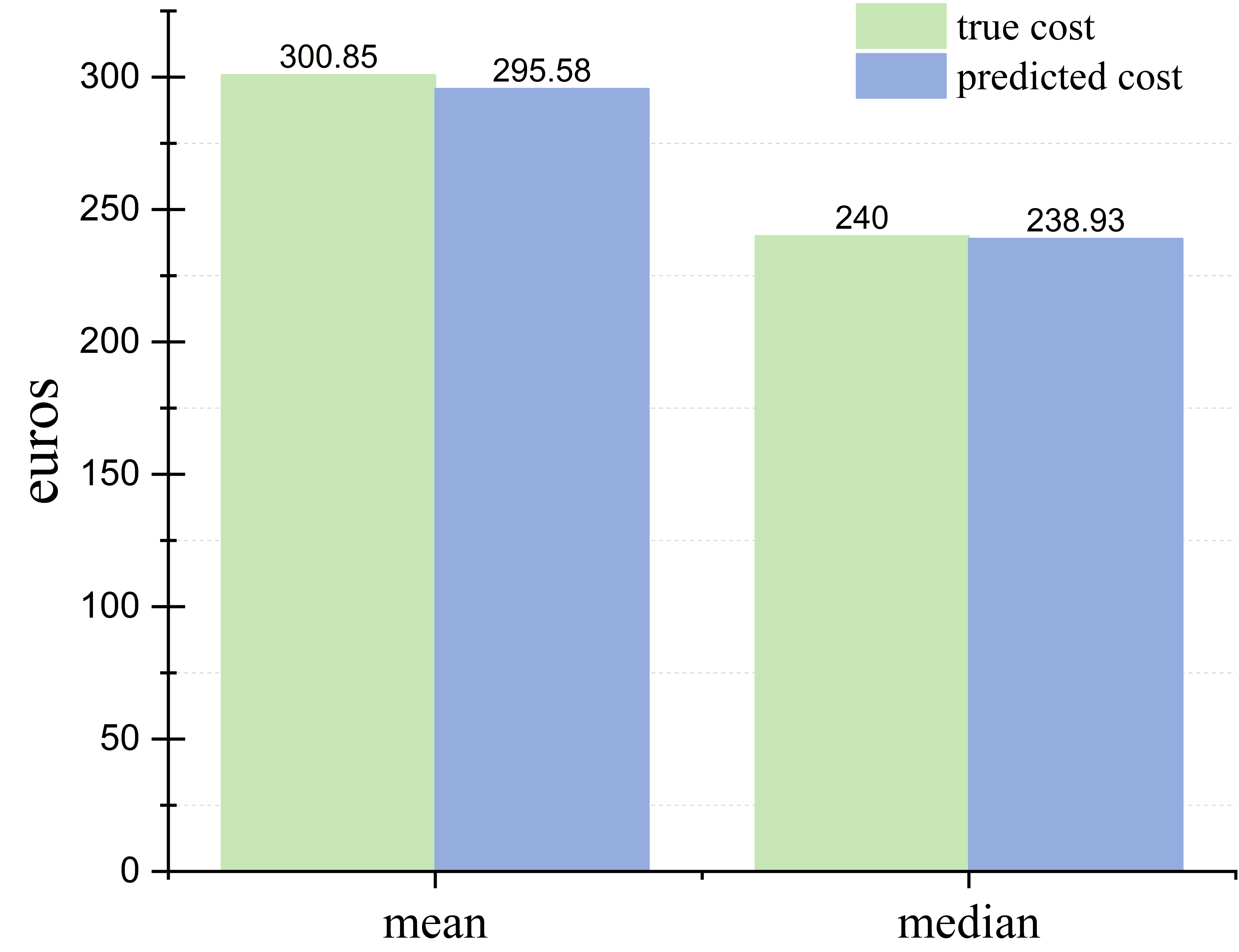}
    \caption{Comparison of mean and median values of true costs and predicted costs. }
    \label{fig_result_barchart}
\end{figure}

Our designed model performs the best based on the results shown in Table \ref{models_results}. Therefore, we only apply this model to predict visit costs and compare them with true visit costs. The sum of the true total cost for all patients in the test set is \euro{146,815}. In comparison, the prediction by the trained model with Transformer augmented data is \euro{144,244}. Fig. \ref{fig_result_barchart} illustrates the comparison of mean and median values.

\section{Discussion and Conclusion} 

In this paper, we have developed a method for predicting (1) the cost types for individual OSA patients' future visits and (2) the total costs resulting from these visits using EHR data.  This method utilizes a versatile Transformer-based architecture which helps make the most of the limited EHR data. While Transformers are applied in many fields -- computer vision, NLP, and also modern healthcare~\cite{lin2022survey}, our approach presents advances relative to existing work by using two Transformers ($\mathcal{M}^{1}$, $\mathcal{M}^{2}$). The two-model system is easy to implement in this context since the encoder portion is identical and enables retaining the sequential patterns of the OSA patients who live in the same region and receive treatment at the same hospital. Model $\mathcal{M}^{1}$ is designed to augment the input by using material that is not suitable as-is for our prediction task. The second component, Transformer $\mathcal{M}^{2}$, outputs not only the prediction of the next year's total costs but also itemization by visit cost type for each visit instance, with the aid of two subsidiary loss functions. Our system outperforms all the baseline models covered in Table \ref{models_results} both in top-\emph{k} accuracy and by regression metrics. We empirically demonstrated prediction improvements arising from our model-informed data augmentation, which enriched the input relative to the original longer-follow-up a third of the high-quality data. We found also that summation of embeddings and utilization of special tokens can serve as an effective way to deal with multivariate sequences of healthcare data. The design sheds new light on approaches to tackle the problem of small bodies of relevant healthcare data and offers a different perspective on cost prediction for better decision-making. 

Data augmentation is one of the best ways to make additional data available for research, and many studies already attest to the effectiveness of a corresponding strategy, via empirical evidence and design insight. However, the difficulty of assessing any given technique in the absence of a large-scale study creates obstacles: quantitative measurement of how well it fits the data is rare, and there is little research into why it works.
Further issues often arise from a lack of variety in the augmented body of data. Frequently, the augmentation is accomplished under supervision, which may result in overfitting or bias in the prediction task~\cite{feng2021survey}. Although a seq2seq model's ability to retain long-term relationships enables it to handle tasks involving lengthy sequences quite well, it is unable to store contextual data. In contrast, the Transformer-based approach preserves data related to context. While it performs all tasks better than seq2seq does, one must bear in mind Transformers' proneness to overfitting with small bodies of data in general, not just in cases of data augmentation~\cite{battaglia2018relational}.

Our future work will consider such issues. We plan to examine other data-augmentation techniques and see how they affect prediction accuracy with datasets of various sizes. Further studies could also extend beyond the healthcare decision-making framework, delving into prediction outcomes at other levels and probing/cultivating direct linkages throughout the systems involved. For example, we should study integration of the cost-prediction model into healthcare's treatment-optimization process. For data specific to OSA, multitasking with EHRs could be handled more effectively. In one example, one might implement mortality/survival evaluation and cost prediction jointly in such a way that not only the financial element but also OSA patients' quality of life would be considered. Since abundant research has shown the power of Bayesian optimization for improved modeling, its use should be considered for informing prediction projects, although the method may prove time-consuming as the models grow larger and more complicated~\cite{cowen2020empirical}.


\section*{Acknowledgment}
We would like to express gratitude to Foundation for Economic Education (grant number 16-9442), the Paulo Foundation, and the Helsinki School of Economics Foundation (HSE). Thanks to Anna Shefl for proofreading the article.

%
%

\vspace*{ 1 cm}
\bibliographystyle{ieeetr}
\bibliography{references}

\begin{thebibliography}{10}

\bibitem{walia2019beyond}
H.~K. Walia, ``Beyond heart health: Consequences of obstructive sleep apnea.,''
  {\em Cleveland Clinic Journal of Medicine}, vol.~86, no.~9 Suppl 1,
  pp.~19--25, 2019.

\bibitem{garbarino2016risk}
S.~Garbarino, O.~Guglielmi, A.~Sanna, G.~L. Mancardi, and N.~Magnavita, ``Risk
  of occupational accidents in workers with obstructive sleep apnea: systematic
  review and meta-analysis,'' {\em Sleep}, vol.~39, no.~6, pp.~1211--1218,
  2016.

\bibitem{bonsignore2019obstructive}
M.~R. Bonsignore, P.~Baiamonte, E.~Mazzuca, A.~Castrogiovanni, and O.~Marrone,
  ``Obstructive sleep apnea and comorbidities: a dangerous liaison,'' {\em
  Multidisciplinary respiratory medicine}, vol.~14, no.~1, pp.~1--12, 2019.

\bibitem{chang2020obstructive}
H.-P. Chang, Y.-F. Chen, and J.-K. Du, ``Obstructive sleep apnea treatment in
  adults,'' {\em The Kaohsiung journal of medical sciences}, vol.~36, no.~1,
  pp.~7--12, 2020.

\bibitem{salman2020obstructive}
L.~A. Salman, R.~Shulman, and J.~B. Cohen, ``Obstructive sleep apnea,
  hypertension, and cardiovascular risk: epidemiology, pathophysiology, and
  management,'' {\em Current Cardiology Reports}, vol.~22, no.~2, pp.~1--9,
  2020.

\bibitem{palomaki2022multimorbidity}
M.~Palom{\"a}ki, T.~Saaresranta, U.~Anttalainen, M.~Partinen, J.~Keto, and
  M.~Linna, ``Multimorbidity and overall comorbidity of sleep apnoea: a finnish
  nationwide study,'' {\em ERJ Open Research}, vol.~8, no.~2, 2022.

\bibitem{sleepapneainFinland2021}
A.~Bachour and H.~Avellan-Hietanen, ``Obstruktiivinen uniapnea aikuisilla
  [obstructive sleep apnea in adults],'' 04 2021.

\bibitem{sleepapneainFinland2019}
A.~U. Saaresranta~Tarja, ``Uniapneaepidemia - mitä hoidolla saavutetaan?
  [sleep apnea epidemics – what do we achieve with treatment?],'' 05 2022.

\bibitem{shah2020secondary}
S.~M. Shah and R.~A. Khan, ``Secondary use of electronic health record:
  Opportunities and challenges,'' {\em IEEE Access}, vol.~8,
  pp.~136947--136965, 2020.

\bibitem{choi2016doctor}
E.~Choi, M.~T. Bahadori, A.~Schuetz, W.~F. Stewart, and J.~Sun, ``Doctor ai:
  Predicting clinical events via recurrent neural networks,'' in {\em Machine
  Learning for Healthcare Conference}, pp.~301--318, PMLR, 2016.

\bibitem{choi2016retain}
E.~Choi, M.~T. Bahadori, J.~Sun, J.~Kulas, A.~Schuetz, and W.~Stewart,
  ``Retain: An interpretable predictive model for healthcare using reverse time
  attention mechanism,'' {\em Advances in Neural Information Processing
  Systems}, vol.~29, 2016.

\bibitem{shickel2017deep}
B.~Shickel, P.~J. Tighe, A.~Bihorac, and P.~Rashidi, ``Deep ehr: a survey of
  recent advances in deep learning techniques for electronic health record
  (ehr) analysis,'' {\em IEEE Journal of Biomedical and Health Informatics},
  vol.~22, no.~5, pp.~1589--1604, 2017.

\bibitem{solares2020deep}
J.~R.~A. Solares, F.~E.~D. Raimondi, Y.~Zhu, F.~Rahimian, D.~Canoy, J.~Tran,
  A.~C.~P. Gomes, A.~H. Payberah, M.~Zottoli, M.~Nazarzadeh, {\em et~al.},
  ``Deep learning for electronic health records: A comparative review of
  multiple deep neural architectures,'' {\em Journal of Biomedical
  Informatics}, vol.~101, p.~103337, 2020.

\bibitem{kennedy2022augmentation}
G.~Kennedy, M.~Dras, and B.~Gallego, ``Augmentation of electronic medical
  record data for deep learning,'' {\em Studies in Health Technology and
  Informatics}, vol.~290, pp.~582--586, 2022.

\bibitem{pramanik2022healthcare}
P.~K.~D. Pramanik, S.~Pal, and M.~Mukhopadhyay, ``Healthcare big data: A
  comprehensive overview,'' {\em Research Anthology on Big Data Analytics,
  Architectures, and Applications}, pp.~119--147, 2022.

\bibitem{o2011impact}
P.~J. O’Connor, J.~M. Sperl-Hillen, W.~A. Rush, P.~E. Johnson, G.~H.
  Amundson, S.~E. Asche, H.~L. Ekstrom, and T.~P. Gilmer, ``Impact of
  electronic health record clinical decision support on diabetes care: a
  randomized trial,'' {\em The Annals of Family Medicine}, vol.~9, no.~1,
  pp.~12--21, 2011.

\bibitem{menachemi2011benefits}
N.~Menachemi and T.~H. Collum, ``Benefits and drawbacks of electronic health
  record systems,'' {\em Risk Management and Healthcare Policy}, vol.~4, p.~47,
  2011.

\bibitem{birkhead2015uses}
G.~S. Birkhead, M.~Klompas, and N.~R. Shah, ``Uses of electronic health records
  for public health surveillance to advance public health,'' {\em Annual Review
  of Public Health}, vol.~36, pp.~345--359, 2015.

\bibitem{wei2019eda}
J.~Wei and K.~Zou, ``Eda: Easy data augmentation techniques for boosting
  performance on text classification tasks,'' {\em arXiv preprint
  arXiv:1901.11196}, 2019.

\bibitem{lu2021textual}
Q.~Lu, D.~Dou, and T.~H. Nguyen, ``Textual data augmentation for patient
  outcomes prediction,'' in {\em 2021 IEEE International Conference on
  Bioinformatics and Biomedicine (BIBM)}, pp.~2817--2821, IEEE, 2021.

\bibitem{csahin2019data}
G.~G. {\c{S}}ahin and M.~Steedman, ``Data augmentation via dependency tree
  morphing for low-resource languages,'' {\em arXiv preprint arXiv:1903.09460},
  2019.

\bibitem{feng2021survey}
S.~Y. Feng, V.~Gangal, J.~Wei, S.~Chandar, S.~Vosoughi, T.~Mitamura, and
  E.~Hovy, ``A survey of data augmentation approaches for nlp,'' {\em arXiv
  preprint arXiv:2105.03075}, 2021.

\bibitem{verma2019manifold}
V.~Verma, A.~Lamb, C.~Beckham, A.~Najafi, I.~Mitliagkas, D.~Lopez-Paz, and
  Y.~Bengio, ``Manifold mixup: Better representations by interpolating hidden
  states,'' in {\em International Conference on Machine Learning},
  pp.~6438--6447, PMLR, 2019.

\bibitem{xie2020unsupervised}
Q.~Xie, Z.~Dai, E.~Hovy, T.~Luong, and Q.~Le, ``Unsupervised data augmentation
  for consistency training,'' {\em Advances in Neural Information Processing
  Systems}, vol.~33, pp.~6256--6268, 2020.

\bibitem{dao2019kernel}
T.~Dao, A.~Gu, A.~Ratner, V.~Smith, C.~De~Sa, and C.~R{\'e}, ``A kernel theory
  of modern data augmentation,'' in {\em International Conference on Machine
  Learning}, pp.~1528--1537, PMLR, 2019.

\bibitem{nie2020named}
Y.~Nie, Y.~Tian, X.~Wan, Y.~Song, and B.~Dai, ``Named entity recognition for
  social media texts with semantic augmentation,'' {\em arXiv preprint
  arXiv:2010.15458}, 2020.

\bibitem{wanyan2021bootstrapping}
T.~Wanyan, J.~Zhang, Y.~Ding, A.~Azad, Z.~Wang, and B.~S. Glicksberg,
  ``Bootstrapping your own positive sample: Contrastive learning with
  electronic health record data,'' {\em arXiv preprint arXiv:2104.02932}, 2021.

\bibitem{perez2018data}
F.~Perez, C.~Vasconcelos, S.~Avila, and E.~Valle, ``Data augmentation for skin
  lesion analysis,'' in {\em OR 2.0 Context-Aware Operating Theaters, Computer
  Assisted Robotic Endoscopy, Clinical Image-Based Procedures, and Skin Image
  Analysis}, pp.~303--311, Springer, 2018.

\bibitem{che2015distilling}
Z.~Che, S.~Purushotham, R.~Khemani, and Y.~Liu, ``Distilling knowledge from
  deep networks with applications to healthcare domain,'' {\em arXiv preprint
  arXiv:1512.03542}, 2015.

\bibitem{choi2016multi}
E.~Choi, M.~T. Bahadori, E.~Searles, C.~Coffey, M.~Thompson, J.~Bost,
  J.~Tejedor-Sojo, and J.~Sun, ``Multi-layer representation learning for
  medical concepts,'' in {\em Proceedings of the 22nd ACM SIGKDD International
  Conference on Knowledge Discovery and Data Mining}, pp.~1495--1504, 2016.

\bibitem{tran2015learning}
T.~Tran, T.~D. Nguyen, D.~Phung, and S.~Venkatesh, ``Learning vector
  representation of medical objects via emr-driven nonnegative restricted
  boltzmann machines (enrbm),'' {\em Journal of Biomedical Informatics},
  vol.~54, pp.~96--105, 2015.

\bibitem{dubois2017effectiveness}
S.~Dubois, N.~Romano, K.~Jung, N.~Shah, and D.~C. Kale, ``The effectiveness of
  transfer learning in electronic health records data,'' {\em ICLR 2017
  Workshop}, 2017.

\bibitem{sutskever2014sequence}
I.~Sutskever, O.~Vinyals, and Q.~V. Le, ``Sequence to sequence learning with
  neural networks,'' {\em Advances in Neural Information Processing Systems},
  vol.~27, 2014.

\bibitem{eddy1996hidden}
S.~R. Eddy, ``Hidden markov models,'' {\em Current opinion in structural
  biology}, vol.~6, no.~3, pp.~361--365, 1996.

\bibitem{lafferty2001conditional}
J.~Lafferty, A.~McCallum, and F.~C. Pereira, ``Conditional random fields:
  Probabilistic models for segmenting and labeling sequence data,'' 2001.

\bibitem{blei2003latent}
D.~M. Blei, A.~Y. Ng, and M.~I. Jordan, ``Latent dirichlet allocation,'' {\em
  Journal of machine Learning research}, vol.~3, no.~Jan, pp.~993--1022, 2003.

\bibitem{dumais2004latent}
S.~T. Dumais {\em et~al.}, ``Latent semantic analysis,'' {\em Annu. Rev. Inf.
  Sci. Technol.}, vol.~38, no.~1, pp.~188--230, 2004.

\bibitem{mikolov2013distributed}
T.~Mikolov, I.~Sutskever, K.~Chen, G.~S. Corrado, and J.~Dean, ``Distributed
  representations of words and phrases and their compositionality,'' {\em
  Advances in Neural Information Processing Systems}, vol.~26, 2013.

\bibitem{mikolov2013efficient}
T.~Mikolov, K.~Chen, G.~Corrado, and J.~Dean, ``Efficient estimation of word
  representations in vector space,'' {\em arXiv preprint arXiv:1301.3781},
  2013.

\bibitem{bojanowski2017enriching}
P.~Bojanowski, E.~Grave, A.~Joulin, and T.~Mikolov, ``Enriching word vectors
  with subword information,'' {\em Transactions of the Association for
  Computational Linguistics}, vol.~5, pp.~135--146, 2017.

\bibitem{pennington2014glove}
J.~Pennington, R.~Socher, and C.~D. Manning, ``Glove: Global vectors for word
  representation,'' in {\em Proceedings of the 2014 Conference on Empirical
  Methods in Natural Language Processing (EMNLP)}, pp.~1532--1543, 2014.

\bibitem{iqbal2020survey}
T.~Iqbal and S.~Qureshi, ``The survey: Text generation models in deep
  learning,'' {\em Journal of King Saud University-Computer and Information
  Sciences}, 2020.

\bibitem{salehinejad2017recent}
H.~Salehinejad, S.~Sankar, J.~Barfett, E.~Colak, and S.~Valaee, ``Recent
  advances in recurrent neural networks,'' {\em arXiv preprint
  arXiv:1801.01078}, 2017.

\bibitem{kingma2013auto}
D.~P. Kingma and M.~Welling, ``Auto-encoding variational bayes,'' {\em arXiv
  preprint arXiv:1312.6114}, 2013.

\bibitem{semeniuta2017hybrid}
S.~Semeniuta, A.~Severyn, and E.~Barth, ``A hybrid convolutional variational
  autoencoder for text generation,'' {\em arXiv preprint arXiv:1702.02390},
  2017.

\bibitem{yang2017improved}
Z.~Yang, Z.~Hu, R.~Salakhutdinov, and T.~Berg-Kirkpatrick, ``Improved
  variational autoencoders for text modeling using dilated convolutions,'' in
  {\em International Conference on Machine Learning}, pp.~3881--3890, PMLR,
  2017.

\bibitem{kingma2019introduction}
D.~P. Kingma, M.~Welling, {\em et~al.}, ``An introduction to variational
  autoencoders,'' {\em Foundations and Trends{\textregistered} in Machine
  Learning}, vol.~12, no.~4, pp.~307--392, 2019.

\bibitem{goodfellow2020generative}
I.~Goodfellow, J.~Pouget-Abadie, M.~Mirza, B.~Xu, D.~Warde-Farley, S.~Ozair,
  A.~Courville, and Y.~Bengio, ``Generative adversarial networks,'' {\em
  Communications of the ACM}, vol.~63, no.~11, pp.~139--144, 2020.

\bibitem{bowman2015generating}
S.~R. Bowman, L.~Vilnis, O.~Vinyals, A.~M. Dai, R.~Jozefowicz, and S.~Bengio,
  ``Generating sentences from a continuous space,'' {\em arXiv preprint
  arXiv:1511.06349}, 2015.

\bibitem{li2017adversarial}
J.~Li, W.~Monroe, T.~Shi, S.~Jean, A.~Ritter, and D.~Jurafsky, ``Adversarial
  learning for neural dialogue generation,'' {\em arXiv preprint
  arXiv:1701.06547}, 2017.

\bibitem{li2016deep}
J.~Li, W.~Monroe, A.~Ritter, M.~Galley, J.~Gao, and D.~Jurafsky, ``Deep
  reinforcement learning for dialogue generation,'' {\em arXiv preprint
  arXiv:1606.01541}, 2016.

\bibitem{shi2018toward}
Z.~Shi, X.~Chen, X.~Qiu, and X.~Huang, ``Toward diverse text generation with
  inverse reinforcement learning,'' {\em arXiv preprint arXiv:1804.11258},
  2018.

\bibitem{vaswani2017attention}
A.~Vaswani, N.~Shazeer, N.~Parmar, J.~Uszkoreit, L.~Jones, A.~N. Gomez,
  {\L}.~Kaiser, and I.~Polosukhin, ``Attention is all you need,'' {\em Advances
  in Neural Information Processing Systems}, vol.~30, 2017.

\bibitem{devlin2018bert}
J.~Devlin, M.-W. Chang, K.~Lee, and K.~Toutanova, ``Bert: Pre-training of deep
  bidirectional transformers for language understanding,'' {\em arXiv preprint
  arXiv:1810.04805}, 2018.

\bibitem{rothe2020leveraging}
S.~Rothe, S.~Narayan, and A.~Severyn, ``Leveraging pre-trained checkpoints for
  sequence generation tasks,'' {\em Transactions of the Association for
  Computational Linguistics}, vol.~8, pp.~264--280, 2020.

\bibitem{floridi2020gpt}
L.~Floridi and M.~Chiriatti, ``Gpt-3: Its nature, scope, limits, and
  consequences,'' {\em Minds and Machines}, vol.~30, no.~4, pp.~681--694, 2020.

\bibitem{chatgpt2022}
OpenAI, ``Chatgpt: Optimizing language models for dialogue,'' 11 2022.

\bibitem{rajkomar2018scalable}
A.~Rajkomar, E.~Oren, K.~Chen, A.~M. Dai, N.~Hajaj, M.~Hardt, P.~J. Liu,
  X.~Liu, J.~Marcus, M.~Sun, {\em et~al.}, ``Scalable and accurate deep
  learning with electronic health records,'' {\em NPJ Digital Medicine},
  vol.~1, no.~1, pp.~1--10, 2018.

\bibitem{esteva2019guide}
A.~Esteva, A.~Robicquet, B.~Ramsundar, V.~Kuleshov, M.~DePristo, K.~Chou,
  C.~Cui, G.~Corrado, S.~Thrun, and J.~Dean, ``A guide to deep learning in
  healthcare,'' {\em Nature medicine}, vol.~25, no.~1, pp.~24--29, 2019.

\bibitem{miotto2018deep}
R.~Miotto, F.~Wang, S.~Wang, X.~Jiang, and J.~T. Dudley, ``Deep learning for
  healthcare: review, opportunities and challenges,'' {\em Briefings in
  bioinformatics}, vol.~19, no.~6, pp.~1236--1246, 2018.

\bibitem{xiao2018opportunities}
C.~Xiao, E.~Choi, and J.~Sun, ``Opportunities and challenges in developing deep
  learning models using electronic health records data: a systematic review,''
  {\em Journal of the American Medical Informatics Association}, vol.~25,
  no.~10, pp.~1419--1428, 2018.

\bibitem{nguyen2016mathtt}
P.~Nguyen, T.~Tran, N.~Wickramasinghe, and S.~Venkatesh, ``Deepr: a
  convolutional net for medical records,'' {\em IEEE Journal of Biomedical and
  Health Informatics}, vol.~21, no.~1, pp.~22--30, 2016.

\bibitem{pham2016deepcare}
T.~Pham, T.~Tran, D.~Phung, and S.~Venkatesh, ``Deepcare: A deep dynamic memory
  model for predictive medicine,'' in {\em Pacific-Asia Conference on Knowledge
  Discovery and Data Mining}, pp.~30--41, Springer, 2016.

\bibitem{gupta2022obesity}
M.~Gupta, T.-L.~T. Phan, H.~T. Bunnell, and R.~Beheshti, ``Obesity prediction
  with ehr data: A deep learning approach with interpretable elements,'' {\em
  ACM Transactions on Computing for Healthcare (HEALTH)}, vol.~3, no.~3,
  pp.~1--19, 2022.

\bibitem{ashfaq2019readmission}
A.~Ashfaq, A.~Sant’Anna, M.~Lingman, and S.~Nowaczyk, ``Readmission
  prediction using deep learning on electronic health records,'' {\em Journal
  of biomedical informatics}, vol.~97, p.~103256, 2019.

\bibitem{tang2021embedding}
R.~Tang, H.~Yao, Z.~Zhu, X.~Sun, G.~Hu, Y.~Li, and G.~Xie, ``Embedding
  electronic health records to learn bert-based models for diagnostic decision
  support,'' in {\em 2021 IEEE 9th International Conference on Healthcare
  Informatics (ICHI)}, pp.~311--319, IEEE, 2021.

\bibitem{choi2017gram}
E.~Choi, M.~T. Bahadori, L.~Song, W.~F. Stewart, and J.~Sun, ``Gram:
  graph-based attention model for healthcare representation learning,'' in {\em
  Proceedings of the 23rd ACM SIGKDD International conference on knowledge
  discovery and data mining}, pp.~787--795, 2017.

\bibitem{tay2020efficient}
Y.~Tay, M.~Dehghani, D.~Bahri, and D.~Metzler, ``Efficient transformers: A
  survey,'' {\em ACM Computing Surveys (CSUR)}, 2020.

\bibitem{li2020behrt}
Y.~Li, S.~Rao, J.~R.~A. Solares, A.~Hassaine, R.~Ramakrishnan, D.~Canoy,
  Y.~Zhu, K.~Rahimi, and G.~Salimi-Khorshidi, ``Behrt: transformer for
  electronic health records,'' {\em Scientific Reports}, vol.~10, no.~1,
  pp.~1--12, 2020.

\bibitem{lin2022survey}
T.~Lin, Y.~Wang, X.~Liu, and X.~Qiu, ``A survey of transformers,'' {\em AI
  Open}, 2022.

\bibitem{miller2011survival}
R.~G. Miller~Jr, {\em Survival analysis}.
\newblock John Wiley \& Sons, 2011.

\bibitem{cremonesi2010performance}
P.~Cremonesi, Y.~Koren, and R.~Turrin, ``Performance of recommender algorithms
  on top-n recommendation tasks,'' in {\em Proceedings of the fourth ACM
  conference on Recommender systems}, pp.~39--46, 2010.

\bibitem{draper1998applied}
N.~R. Draper and H.~Smith, {\em Applied regression analysis}, vol.~326.
\newblock John Wiley \& Sons, 1998.

\bibitem{botchkarev2018performance}
A.~Botchkarev, ``Performance metrics (error measures) in machine learning
  regression, forecasting and prognostics: Properties and typology,'' {\em
  arXiv preprint arXiv:1809.03006}, 2018.

\bibitem{ottom2022znet}
M.~A. Ottom, H.~A. Rahman, and I.~D. Dinov, ``Znet: Deep learning approach for
  2d mri brain tumor segmentation,'' {\em IEEE Journal of Translational
  Engineering in Health and Medicine}, 2022.

\bibitem{yuan2020using}
Y.~Yuan, L.~Lin, L.-Z. Huo, Y.-L. Kong, Z.-G. Zhou, B.~Wu, and Y.~Jia, ``Using
  an attention-based lstm encoder--decoder network for near real-time
  disturbance detection,'' {\em IEEE Journal of Selected Topics in Applied
  Earth Observations and Remote Sensing}, vol.~13, pp.~1819--1832, 2020.

\bibitem{battaglia2018relational}
P.~W. Battaglia, J.~B. Hamrick, V.~Bapst, A.~Sanchez-Gonzalez, V.~Zambaldi,
  M.~Malinowski, A.~Tacchetti, D.~Raposo, A.~Santoro, R.~Faulkner, {\em
  et~al.}, ``Relational inductive biases, deep learning, and graph networks,''
  {\em arXiv preprint arXiv:1806.01261}, 2018.

\bibitem{cowen2020empirical}
A.~I. Cowen-Rivers, W.~Lyu, R.~Tutunov, Z.~Wang, A.~Grosnit, R.~R. Griffiths,
  H.~Jianye, J.~Wang, and H.~B. Ammar, ``An empirical study of assumptions in
  bayesian optimisation,'' {\em arXiv preprint arXiv:2012.03826}, vol.~445,
  2020.

\end{thebibliography}
%

%
%
%
%
%
%
%
\end{document}